\documentclass{article} 
\usepackage{iclr2025_conference,times}

\usepackage{amsmath,amsfonts,bm}

\def\eqref#1{equation~\ref{#1}}

\def\1{\bm{1}}

\DeclareMathAlphabet{\mathsfit}{\encodingdefault}{\sfdefault}{m}{sl}
\SetMathAlphabet{\mathsfit}{bold}{\encodingdefault}{\sfdefault}{bx}{n}

\usepackage{amsmath}
\usepackage{booktabs}
\usepackage{graphicx}
\usepackage{url}
\usepackage{float}
\usepackage{multicol}
\usepackage{xcolor}
\usepackage{enumitem}
\usepackage{hyperref}

\title{\fontsize{16pt}{20pt}\selectfont LoongSage: A Production-Grade, High-Efficiency Agentic RL Framework for Frontier LLMs with Validated Recipes}

\author{
    \vspace{80pt}
    \makebox[\textwidth][c]{
        \hspace*{-2.8em} 
        \textbf{\Large \raisebox{-0.4\height}{\includegraphics[height=2.5em]{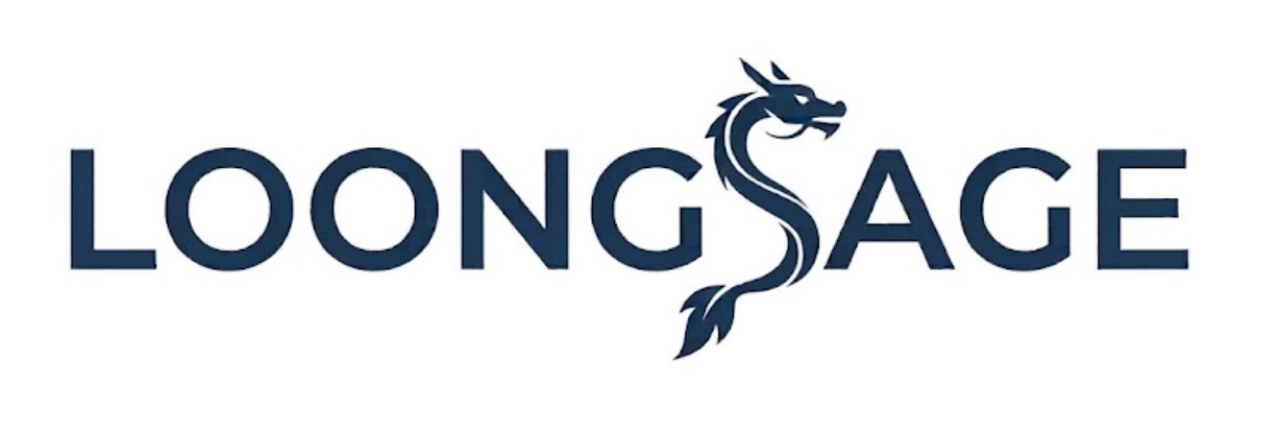}}~Team}%
    }
}

\iclrfinalcopy

\begin{document}

\maketitle

\vspace{-112pt}
\begin{center}
       \href{https://github.com/baidu-baige/LoongSage/}{%
        \raisebox{-0.15\height}{\includegraphics[height=1.2em]{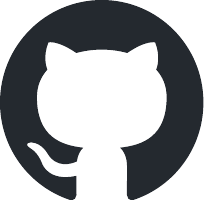}}~%
        \texttt{https://github.com/baidu-baige/LoongSage}%
    }
\end{center}

\vspace{1em}
\begin{abstract}
Reinforcement Learning (RL) has emerged as a critical phase in the post-training pipeline of Large Language Models (LLMs). As the frontier of LLM capabilities shifts from static reasoning to agentic tasks---characterized by external tool invocation, multi-turn interactions, and long-horizon decision-making---existing mainstream RL frameworks have grown increasingly unwieldy, raising the overhead of adoption, usage, and debugging.
To address these challenges, we present LoongSage, an agent-native, fully asynchronous RL framework with a lightweight core of roughly 16K lines of code. Built upon Megatron for distributed training and SGLang for high-throughput inference, LoongSage centralizes orchestration within a single controller process and encapsulates all GPU-resident components as Ray actors. It introduces AgentFlow, a unified execution paradigm that orchestrates both white-box and black-box agents. For asynchronous execution, it integrates sliding-window strategies that place configurable bounds on trajectory staleness, and employs TransferMesh, a topology-aware synchronization mechanism that routes weights after each step over the fastest transport available to each receiver.
Beyond agentic RL, LoongSage incorporates a memory-efficient Teacher Manager that decouples the scaling of teacher models from GPU memory constraints, natively supporting Multi-Teacher On-Policy Distillation. Four complete training runs, spanning RL and multi-teacher distillation, show rising task rewards in the RL runs, with their train--inference gap held at the $10^{-4}$ scale, and a controlled ablation shows that the fully asynchronous pipeline achieves up to a $1.40\times$ steady-state speedup over a colocated baseline.
\end{abstract}

\section{Introduction}
\label{sec:introduction}

Reinforcement Learning (RL) has emerged as a pivotal post-training mechanism driving recent capability gains in Large Language Models (LLMs). Historically, the GPT series leveraged Reinforcement Learning from Human Feedback (RLHF) to translate human preferences into optimizable objectives, establishing the viability of instruction-following assistants~\citep{instructgpt}. More recently, models like DeepSeek-R1 have demonstrated that, given verifiable rewards, complex behaviors such as extended chain-of-thought reasoning, self-correction, and backtracking can emerge through RL optimization~\citep{r1,chainofthought}. As the focus transitions from instruction following to complex reasoning, RL has shifted from a secondary fine-tuning step to a central phase for instilling core model capabilities.

Despite these reasoning advancements, text generation alone is insufficient for executing complex, real-world projects. This limitation has catalyzed exploration into tool-calling capabilities~\citep{react,toolformer}, driving the paradigm shift towards agentic RL. In this context, reward signals are no longer confined to the correctness of single-turn responses. Instead, they evaluate an agent's holistic ability to complete tasks by invoking tools, manipulating files, executing tests, and refining actions iteratively based on environmental feedback. Consequently, coding, research, and computer-use agents have established end-to-end task completion as a new capability frontier~\citep{sweagent,webarena,osworld}. Compared to single-turn tasks, agentic RL unfolds over extended horizons with persistent multi-turn states, necessitates sandboxed execution for external side effects, and produces heavy-tailed episode distributions~\citep{areal,agentlightning}. This fundamental shift redefines the role of an RL framework from merely generating and scoring text to continuously driving and recording a system operating within dynamic environments.

However, many existing RL frameworks are rooted in assumptions inherited from single-turn RLHF or verifiable-reward workloads~\citep{openrlhf}, rendering them suboptimal for agentic training. While some systems introduce controller-centered abstractions~\citep{verl} or deliberately narrow the pairing of backend engines~\citep{slime}, few treat the complexities of agentic training as first-class design concerns, and these complexities translate directly into production overheads. For instance, memory constraints typically bottleneck the number of teachers in multi-teacher distillation, and synchronous train--inference loops suffer severe utilization drops due to long-tailed trajectory rollouts. Furthermore, several systemic discrepancies manifest as implicit off-policy drift: mismatches in tokenization, numerical precision, and Mixture-of-Experts (MoE) routing between the training and inference engines. These are compounded by loose trajectory reconstruction---a subtle but accumulating divergence between the inference-generated sequences and the training-consumed data. Because algorithmic hyperparameters cannot tune away these systemic issues, a modern framework must elevate resource management, train--inference decoupling, numerical consistency, and trajectory fidelity to first-class design primitives.

To address these critical gaps, we present \emph{LoongSage}, a lightweight agentic RL training framework engineered for production-grade workloads. It is built on Megatron~\citep{megatron} for distributed training and SGLang~\citep{sglang} for high-throughput inference. Notably, we open-source not only the framework but also the validated training recipes, enabling researchers to initialize runs from configurations proven on frontier models rather than starting from scratch. Its key features include:

\begin{itemize}
\item \textbf{Seamless Integration of Any Agent:} LoongSage accommodates any white-box or black-box agent via standard LLM serving APIs (e.g., Chat Completions) with zero intrusion into agent logic. Its AgentFlow paradigm maintains strict token-level alignment (Token-In, Token-Out; TITO), feeding the training engine the exact token streams generated during inference.
\item \textbf{Staleness-Controlled Fully Asynchronous Training:} By decoupling the sampling and optimization phases via a train--rollout buffer, the fully asynchronous pipeline overlaps rollout with training and shortens the step time. To control the resulting trajectory staleness, it employs sliding-window strategies, keeping off-policy drift explicit and observable.
\item \textbf{Topology-Aware Weight Synchronization:} Through TransferMesh, LoongSage transfers weights from training to inference as a continuous named tensor stream. The system selects the transport by device topology---a CUDA IPC handle for colocated GPUs or an NCCL broadcast for other receivers---keeping the synchronization process transparent to both backend engines.
\item \textbf{Full-Vocabulary Multi-Teacher On-Policy Distillation:} An independent Teacher Manager treats teacher models as dynamically managed compute resources, decoupling the permissible number of teachers from the student model's GPU memory constraints to enable scalable multi-teacher distillation. Furthermore, it natively supports a composable, multi-paradigm objective function---spanning PG-style, GKD-style, and hybrid distillation modes---equipped with built-in estimators including sampled-token, Top-$K$, full-vocabulary, and Jensen--Shannon Divergence (JSD) KL strategies.
\item \textbf{Systemic Train--Inference Consistency:} LoongSage combines Deterministic Source Alignment (e.g., MoE routing replay), selective precision pinning for critical layers, and algorithmic residual corrections (including TIS, M2PO, and OPSM) to keep the optimization plane tracking the sampling distribution.
\item \textbf{Lightweight, Usable, and Highly Extensible:} By building on pinned upstream versions of its backend engines rather than forks, the lightweight core inherits upstream optimizations and newly supported models with minimal adaptation. Additionally, a highly extensible plugin system exposes ten core extension points---spanning agents, reward models, and scheduling strategies---so that new algorithms can be plugged in through configuration without modifying the framework's main trunk.
\end{itemize}
\section{Related Work}
\label{sec:related}

\subsection{The Evolution of LLM Post-Training}

\paragraph{Compositional Paradigms.}
Modern LLM post-training is converging toward a compositional pipeline rather than a single optimization objective. Supervised fine-tuning establishes base capabilities and response formats, after which Reinforcement Learning optimizes the policy against a reward signal. 
Building on knowledge distillation~\citep{distillation}, on-policy distillation then adds a denser supervision channel, transferring the output distributions of one or more teachers onto the student's own rollouts~\citep{minillm,gkd}. 
These stages differ in where supervision originates and in whether generation happens inside the training loop, yet they increasingly share the same training and inference infrastructure. 
Frameworks therefore need to compose objectives and data flows without forcing every new method into a separate execution stack.

\paragraph{From RLHF to RLVR.}
Reinforcement Learning entered LLM post-training as RLHF, which fits a reward model to preference comparisons and optimizes the policy with PPO~\citep{ppo,instructgpt}. Later work then moved in two directions, one of which removes online RL altogether by reparameterizing the preference objective, as DPO does~\citep{dpo}. The other retains online RL but replaces the learned reward model with a programmatic checker, which in turn simplifies the algorithms themselves. GRPO drops the value network in favor of group-relative advantages~\citep{deepseekmath}, and DAPO combines decoupled clipping, dynamic sampling, and token-level loss aggregation for large-scale RLVR~\citep{dapo}. The transition also moves the systems cost from reward-model and reference-policy management toward rollout generation, program execution, sample filtering, and reliable aggregation of completed outcomes.

\paragraph{Agentic RL.}
As models are deployed in more complex settings, RL is shifting toward agentic tasks with multi-turn interaction and external environments, building on early evidence that LLMs can be taught to invoke tools and APIs, whether by in-context prompting or by fine-tuning~\citep{react,toolformer,toolllm}. An agentic reward evaluates whether a task was completed rather than whether a single response was correct, so producing it depends on tool use and sandboxed execution against a live environment~\citep{sweagent,swebench,osworld}. Unlike single-turn generation, the resulting trajectory contains intermediate observations, actions, and tool results whose contribution to the final outcome may be delayed. A practical training system must therefore preserve the generated sequence and its metadata across multiple turns, tolerate heavy-tailed episode lengths and external-environment failures, and manage the sandbox lifecycle~\citep{agentlightning}.

\subsection{Distributed Training and Inference Systems}

\paragraph{Parallelism and Distributed Training.}
Large-model training combines data, tensor, pipeline, and expert parallelism to overcome single-GPU limits, and Megatron-LM provides optimized implementations of these strategies~\citep{megatron}. DeepSpeed and ZeRO instead reduce memory use by partitioning optimizer states, gradients, and parameters~\citep{deepspeed,zero}. On top of these placement decisions, RL post-training adds a further requirement, because the inference side must generate rollouts concurrently or alternately with the distributed training side. Resource placement and the synchronization relation between the two sides therefore become part of the framework design.

\paragraph{Orchestration and Inference Engines.}
Ray manages heterogeneous pipeline components across a cluster~\citep{ray}, while rollout generation relies on dedicated serving engines. vLLM improves throughput through continuous batching and PagedAttention~\citep{vllm}, and SGLang targets multi-turn and structured LLM programs~\citep{sglang}. Because rollout requests have variable lengths and may invoke tools whereas training requires tightly synchronized collective computation, orchestration must coordinate two distinct lifecycles and resource layouts rather than treating inference as a simple preprocessing step.

\subsection{RL Frameworks}

\paragraph{Foundational RL Frameworks.}
OpenRLHF~\citep{openrlhf} established an accessible combination of distributed training and high-throughput inference. NeMo-Aligner~\citep{nemoaligner} targets Megatron-based scale, TRL~\citep{trl} emphasizes Hugging Face usability, verl~\citep{verl} uses a controller-centered HybridFlow model, and slime~\citep{slime} narrows the dependency surface by pairing Megatron-LM with SGLang. Together these systems span accessibility, ecosystem integration, large-scale backend support, expressive control flow, and a deliberately small dependency surface. Agentic workloads, however, further require explicit support for trajectory structure, resource sharing, and the boundary between rollout and optimization.

\paragraph{Asynchronous Execution and Rollout Management.}
AReaL~\citep{areal} makes asynchronous generation and training the default and mitigates long-tail latency by buffering unfinished trajectories for later continuation, which improves utilization at the cost of trajectory staleness. Agent-Lightning~\citep{agentlightning} instead disaggregates the agent from the trainer, while agent frameworks and benchmarks such as SWE-agent~\citep{sweagent}, OpenHands~\citep{openhands}, WebArena~\citep{webarena}, and OSWorld~\citep{osworld} motivate support for long-horizon interaction and persistent state.
These directions expose a central trade-off in which decoupling improves utilization and accommodates long-running interactions while also increasing coordination complexity and the possibility that training consumes data produced by older policy versions. A framework should therefore expose this trade-off as a configurable policy rather than a fixed behavior, and keep one interface for trajectories, weight transfer, and training regardless of the execution mode.

\section{The LoongSage Framework}
\label{sec:framework}

\begin{figure}[htbp]
\centering
\includegraphics[width=\linewidth]{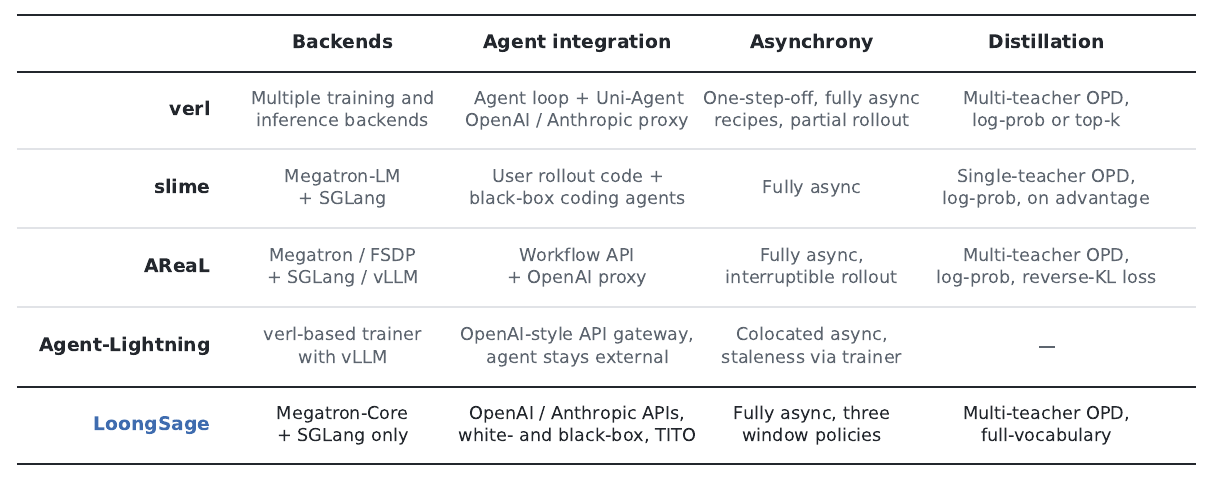}
\caption{Positioning of LoongSage relative to closely related systems. Each entry is compiled from the publicly available documentation of the corresponding project at the time of writing. A dash indicates a capability that is not documented, rather than one that cannot be realized within the system.}
\label{fig:framework-comparison}
\end{figure}

This section describes the system architecture of LoongSage and the orchestration of its principal components within an agentic RL training pipeline. By establishing strict boundaries across orchestration, data representation, and backend integration, the framework accommodates both synchronous and fully asynchronous execution within a unified architecture. Figure~\ref{fig:framework-comparison} situates LoongSage against the most closely related systems across four structural axes: training and inference backends, agent integration methodology, support for multi-teacher distillation, and execution asynchrony. Rather than pursuing broad backend agnosticism or unrestricted scheduling, LoongSage commits to an optimized pairing of specific engines and staleness-controlled asynchronous execution. This focus lets the framework address the central challenges of agentic workloads: ensuring the fidelity of trajectories fed to the optimizer, orchestrating multi-teacher distillation at scale, and maintaining observability over the train--inference gap.

\subsection{Design Trade-offs}

The architectural design of LoongSage is guided by three core principles.

\textbf{One controller, many workers.} All system orchestration is centralized within a single controller process, where the \texttt{Trainer} acts as the sole state machine dictating every phase transition. GPU-resident components operate as Ray actors, exposing coarse-grained control methods that return \texttt{ObjectRef}s. Keeping calls coarse amortizes the overhead of Ray serialization and remote procedure calls (RPCs). Crucially, the controller executes no forward passes and holds no model parameters, guaranteeing that the global state of a long-running distributed job can be fully inspected from this single process.

\textbf{Token-level internal boundaries.} The internal interface between the sampling and optimization planes operates strictly at the token level---exchanging token sequences coupled with their respective log-probabilities, loss masks, and weight versions. Human-readable text exists only at the outermost protocol edge where the agent interacts. This abstraction eliminates re-tokenization discrepancies, a major source of implicit train--inference divergence (Section~\ref{sec:agentflow}).

\textbf{Thin trunk with named seams.} LoongSage builds natively upon Megatron-Core, Megatron-Bridge, and SGLang without introducing cumbersome abstraction layers for alternative backends. To maximize extensibility, algorithmic components---such as agents, reward models, policy losses, and asynchronous scheduling strategies---are exposed via named registry entries (seams) rather than through rigid subclassing of a monolithic framework class.

\subsection{System Organization}

LoongSage structurally decomposes the RL pipeline into seven interacting planes: control, sampling, training, algorithm, data, transfer, and resource. These planes communicate exclusively via well-defined data objects and control interfaces.

The control plane is anchored by the \texttt{Trainer}, which orchestrates the state transitions for each algorithmic step. Dedicated lifecycle managers---\texttt{TrainManager}, \texttt{RolloutManager}, and \texttt{TeacherManager}---oversee the training, inference, and teacher worker pools, respectively. The \texttt{RolloutSampler} maps step-level data requirements into batched sampling tasks. To prevent tangled dependencies, these managers never invoke one another directly. Instead, the \texttt{Trainer} sequences them according to the current training phase.

\begin{figure}[htbp]
\centering
\includegraphics[width=.95\linewidth]{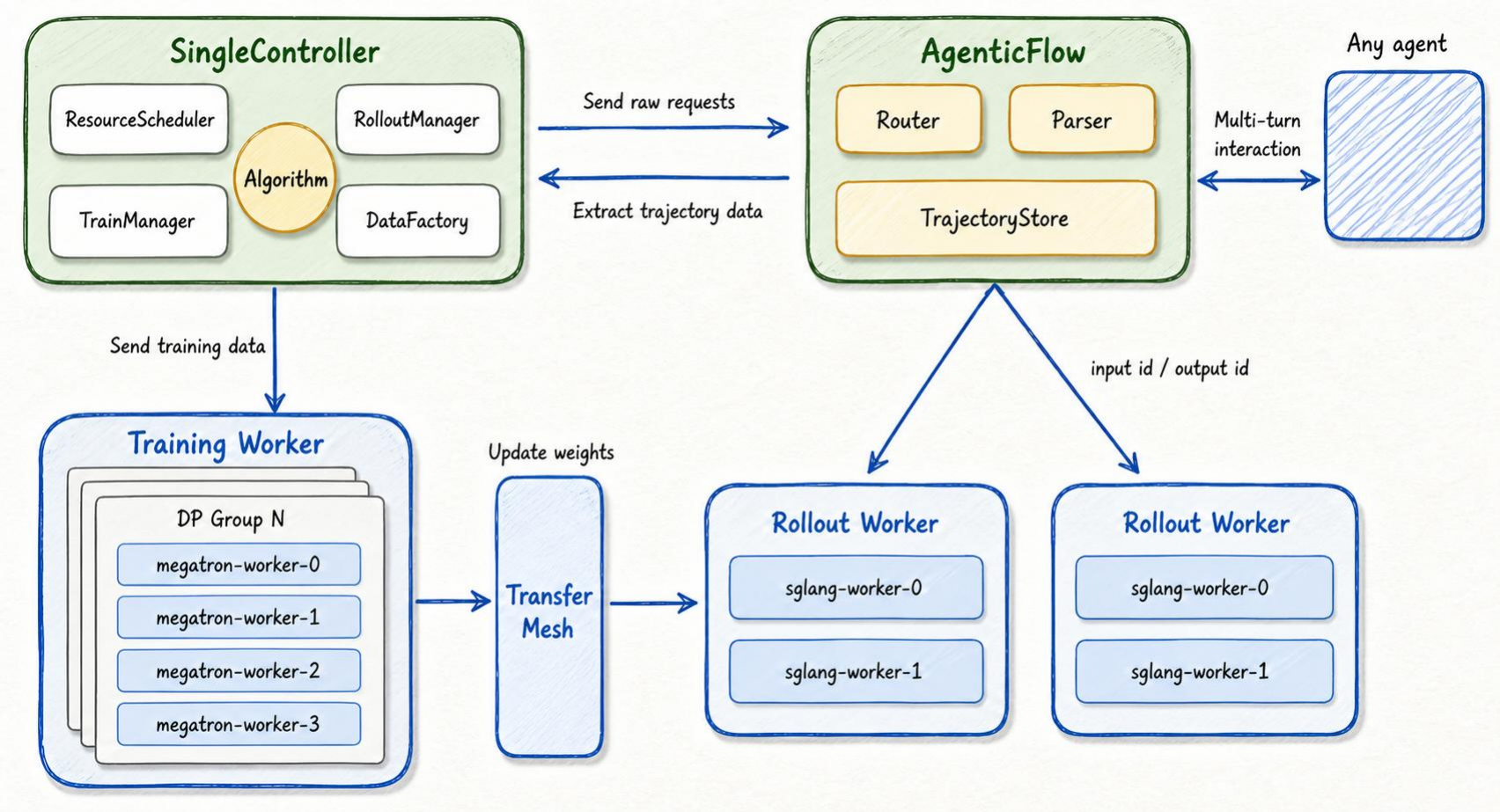}
\caption{Overall system architecture of LoongSage.}
\label{fig:architecture}
\end{figure}

As illustrated in Figure~\ref{fig:architecture}, the sampling plane is governed by AgentFlow, which translates unstructured multi-turn agent interactions into trainable trajectories. By operating the agent externally, LoongSage embraces the disaggregated design philosophy advocated by Agent-Lightning. AgentFlow's token-level alignment and multi-agent support are detailed in Section~\ref{sec:agentflow}. The training plane wraps the Megatron-based worker pool that executes the forward, backward, and optimizer passes over admitted trajectories. The algorithm plane exposes registries for advantage estimation (e.g., GRPO~\citep{deepseekmath} and GAE~\citep{gae}, which requires a value model trained on a separate Megatron critic worker colocated with the actor), policy losses (e.g., the PPO clipped surrogate~\citep{ppo} and GSPO~\citep{gspo}), and KL divergence constraints, and additionally provides configurable importance-sampling corrections~\citep{tis,offpolicyrlframework} and off-policy sequence masking~\citep{deepseekv3.2}.

The data plane governs Multi-Data-Source Mixed Training, managing stateful data loading, admission filtering, and data-parallel partitioning. It enables the independent configuration of agents, rewards, and sampling parameters per source, while strictly maintaining global data mixture proportions invariant across all ranks and mini-batches. The transfer plane, instantiated as TransferMesh (Section~\ref{sec:transfermesh}), manages topology-aware weight synchronization. Finally, the resource plane orchestrates cluster allocations via Ray placement groups and GPU bundles.

\subsection{Training Flow and Placement}

Under the default synchronous execution mode, a training step proceeds through four distinct stages. During the \emph{sampling stage}, the \texttt{RolloutSampler} fetches prompt groups from various data sources and dispatches sampling tasks. In the \emph{admission stage}, completed trajectories are grouped and evaluated by a data filter. Because rejections occur post-generation, the sampler proactively over-samples a configurable buffer per data source, refilling rejected slots within a bounded budget and enforcing step-level timeouts for stalled requests. During the \emph{aggregation stage}, the controller aligns admitted trajectories with the active policy version and data-parallel layout, while teacher workers compute necessary distillation signals. Finally, in the \emph{update stage}, training workers execute the forward, backward, and optimizer passes, concluding with the transfer of updated parameters back to the inference engine for the subsequent rollout.

The fully asynchronous mode eliminates this strict synchronization barrier, enabling the sampling and optimization engines to make concurrent progress via a decoupled train--rollout buffer. The buffering, freshness constraints, and recovery mechanics of this mode are detailed in Section~\ref{sec:async}. 

\begin{figure}[htbp]
\centering
\includegraphics[width=0.8\linewidth]{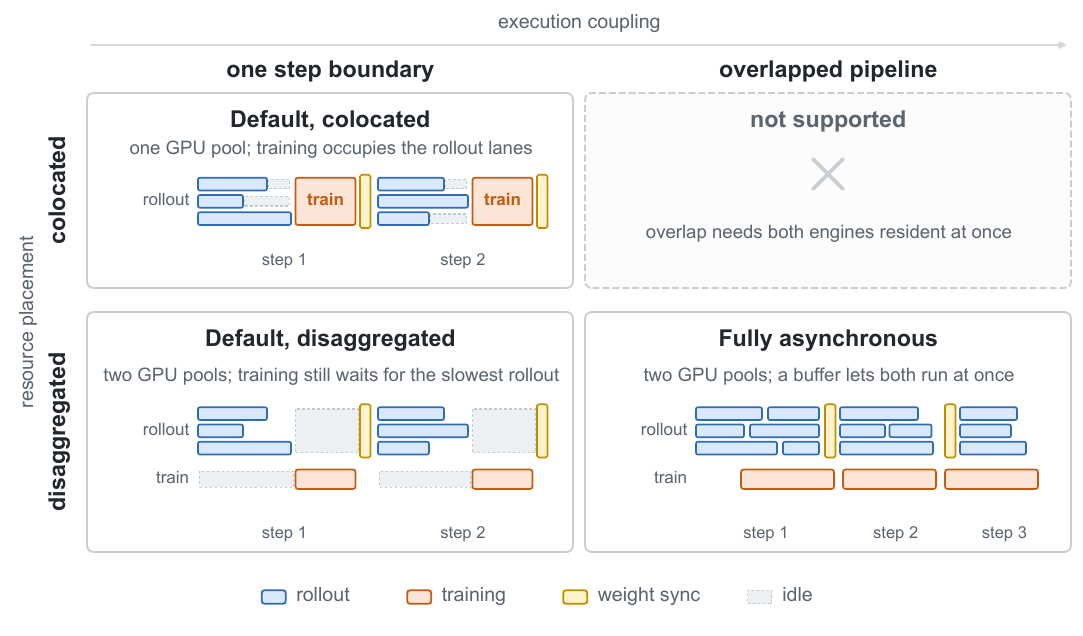}
\caption{The two orthogonal configuration axes behind LoongSage's operational modes, their trade-offs, and the combination that the framework does not support.}
\label{fig:execution-modes}
\end{figure}

Execution coupling and resource placement are treated as orthogonal architectural axes (Figure~\ref{fig:execution-modes}). \emph{Colocated placement} time-multiplexes device memory so that both engines and the weight-update phase fit on the same GPUs, whereas \emph{disaggregated placement} dedicates disjoint hardware pools to each engine. Both placement strategies share an identical underlying scheduler, controller, AgentFlow paradigm, and weight-transfer interface. Among the four possible combinations of these two axes, three are structurally supported. LoongSage does not support the fully asynchronous + colocated combination, since overlapping rollout and training requires both engines resident in device memory at once. That said, the colocated asynchronous mode of Agent-Lightning v1.0~\citep{agentlightningv1} listed in Figure~\ref{fig:framework-comparison} is a different notion: rollout and training still alternate on a shared GPU pool, and the asynchrony comes from the gateway draining in-flight LLM requests at each update and resuming unfinished rollouts under the new weights afterward. This is closer to LoongSage's colocated mode with partial rollout than to concurrent overlap.

\section{Key Features}
\label{sec:features}

\subsection{Seamless Integration of Any Agent}
\label{sec:agentflow}

\begin{figure}[htbp]
\centering
\includegraphics[width=0.85\linewidth]{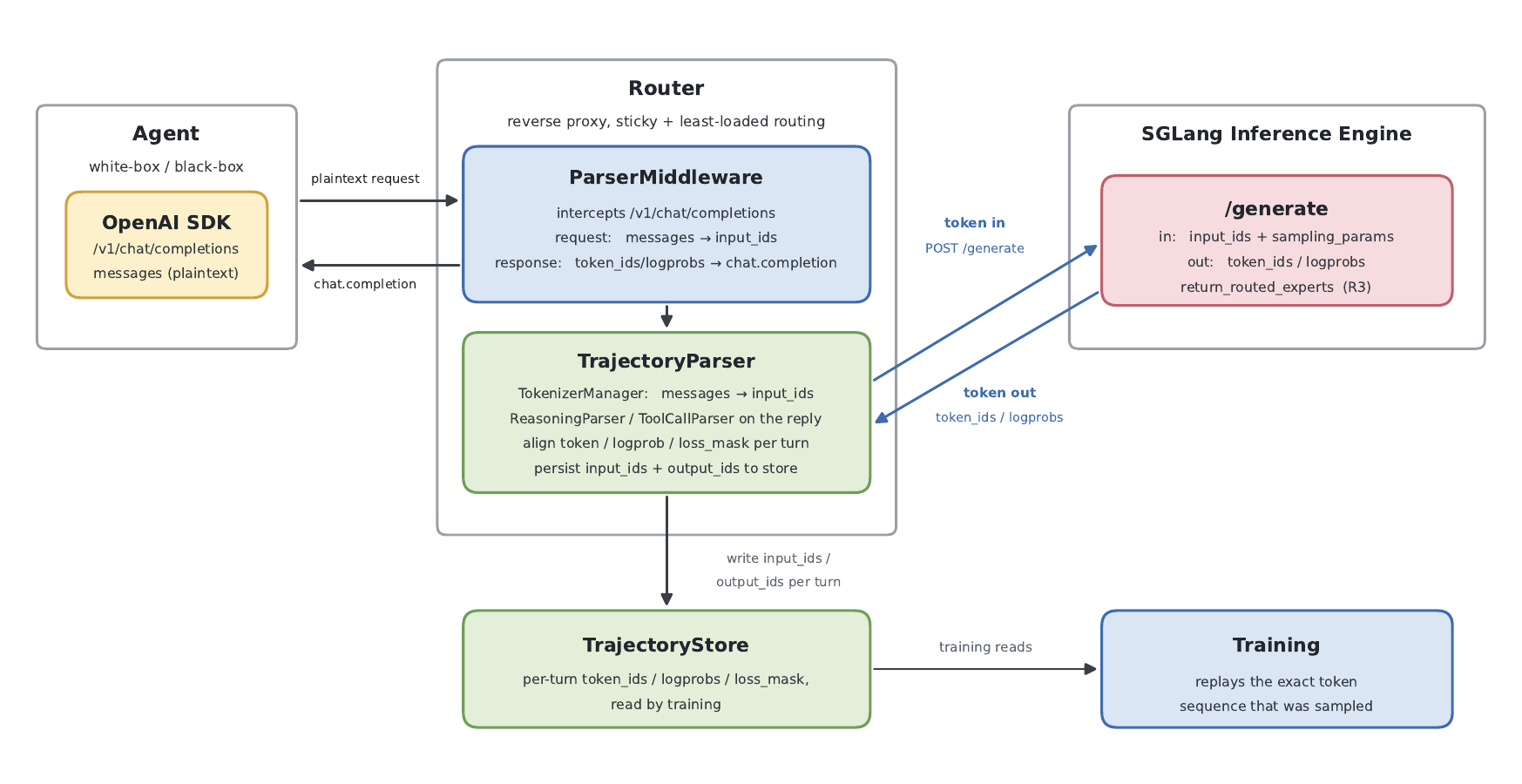}
\caption{The AgentFlow request path. The Router intercepts each Chat Completions request, tokenizes the turn, dispatches it to SGLang's \texttt{/generate}, and records the returned tokens per turn in the \texttt{TrajectoryStore}. Other protocols are shown in Figure~\ref{fig:agent-adapter}.}
\label{fig:seamless-integration}
\end{figure}

LoongSage integrates arbitrary agents with zero intrusion into their internal logic. Agents interact with the framework exclusively via a per-trajectory router URL exposing standard LLM serving APIs, remaining entirely agnostic to the underlying tokenization, trajectory storage, and sandbox lifecycles. The router serves OpenAI Chat Completions by default, and protocol adapters bridge OpenAI Responses and Anthropic Messages. AgentFlow operates as the sampling plane enabling this abstraction---mediating among agents, the inference engine, and the training data pipeline to convert multi-turn interactions into replayable RL trajectories, as illustrated in Figure~\ref{fig:seamless-integration}.

Supported agents fall into two structurally distinct categories. A \emph{white-box agent} implements its interaction loop internally within LoongSage, directly controlling tool invocations and LLM requests. Consequently, its trajectory forms a linear, append-only sequence of prompts, responses, tool execution results, and termination signals. Conversely, a \emph{black-box agent} executes as an opaque binary or script within an isolated sandbox. Here, the framework manages the entry point and intercepts outgoing requests, yielding complex, non-linear trajectories that require a tree-structured storage backend. The current repository natively supports white-box GSM8K~\citep{gsm8k}, BrowseComp-Plus~\citep{browsecompplus}, and Mini-SWE agents, alongside black-box OpenCode~\citep{opencode}, Codex, and Claude Code agents.

\begin{figure}[htbp]
\centering
\includegraphics[width=\linewidth]{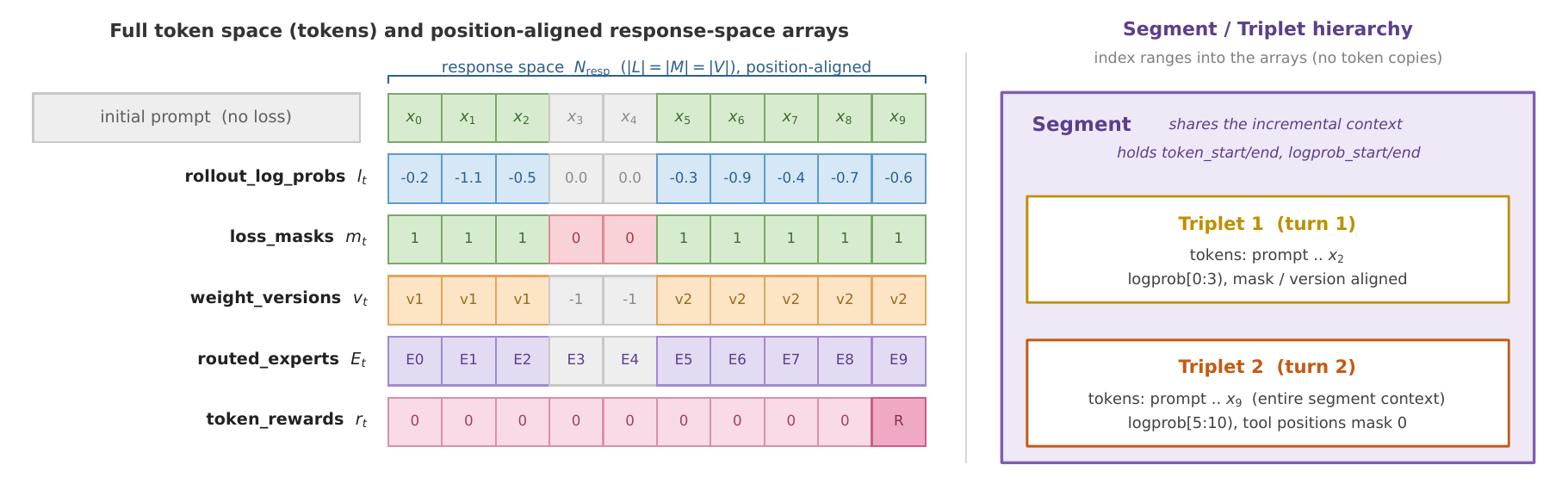}
\caption{Data representation under TITO: a single flattened token array with its position-aligned response-space arrays, referenced by \texttt{Segment} and \texttt{Triplet} indices rather than copied. For readability, the routed-expert row is drawn over the response span only; in the code it is indexed over the full token sequence.}
\label{fig:feature-agentflow}
\end{figure}

As depicted in Figure~\ref{fig:feature-agentflow}, AgentFlow represents each trajectory as a flattened, contiguous token array, tightly coupled with position-aligned response-space arrays for loss masks, rollout log-probabilities, weight versions, and token-level rewards. When Rollout Routing Replay (R3; Section~\ref{sec:consistency}) is enabled, a routed-expert map additionally records the inference engine's routing for every token position, tool tokens included. To minimize memory overhead, a \texttt{Segment} struct tracks token and response spans, while a \texttt{Triplet} struct records individual model interactions using index ranges rather than duplicating token data. Tool observations are injected into the response space with a zeroed loss mask, placeholder log-probabilities, and an invalid ($-1$) weight version. Generated tokens carry the weight version of the policy that produced them, so a trajectory that spans an asynchronous model update records the version change explicitly. Finally, only the terminal position encodes the scalar task reward $R$, ensuring all sampling statistics for generated tokens remain intact throughout the pipeline.

\textbf{Token-In, Token-Out (TITO).} A core mechanism of AgentFlow is storing the exact tokens consumed and generated by the inference engine, deliberately avoiding the practice of re-tokenizing textual messages prior to training. Such re-tokenization often introduces compounding discrepancies between the sampled and trained sequences due to whitespace handling, special tokens, reasoning markers, or tool-call serialization. Under TITO, high-level textual requests and responses are mapped transparently to their internal token representations at the framework's outermost boundary, ensuring the sampled and trained sequences remain identical. 

Whenever a request extends the existing conversation history, AgentFlow reuses the established token prefix and tokenizes only the new messages. When an agent (typically a black-box one) rewrites its context, AgentFlow opens a new segment and re-tokenizes the mutated context while preserving its topological relationship to earlier segments. Should a long-horizon trajectory be interrupted, its sandbox state can also be preserved and reused, preventing the loss of intermediate file edits or process states. The framework's sandbox interface provides native Kubernetes and Docker backends. Thus, agents requiring shell execution run inside ephemeral pods or containers whose lifecycles are strictly bound to the trajectory.

\begin{figure}[htbp]
\centering
\includegraphics[width=\linewidth]{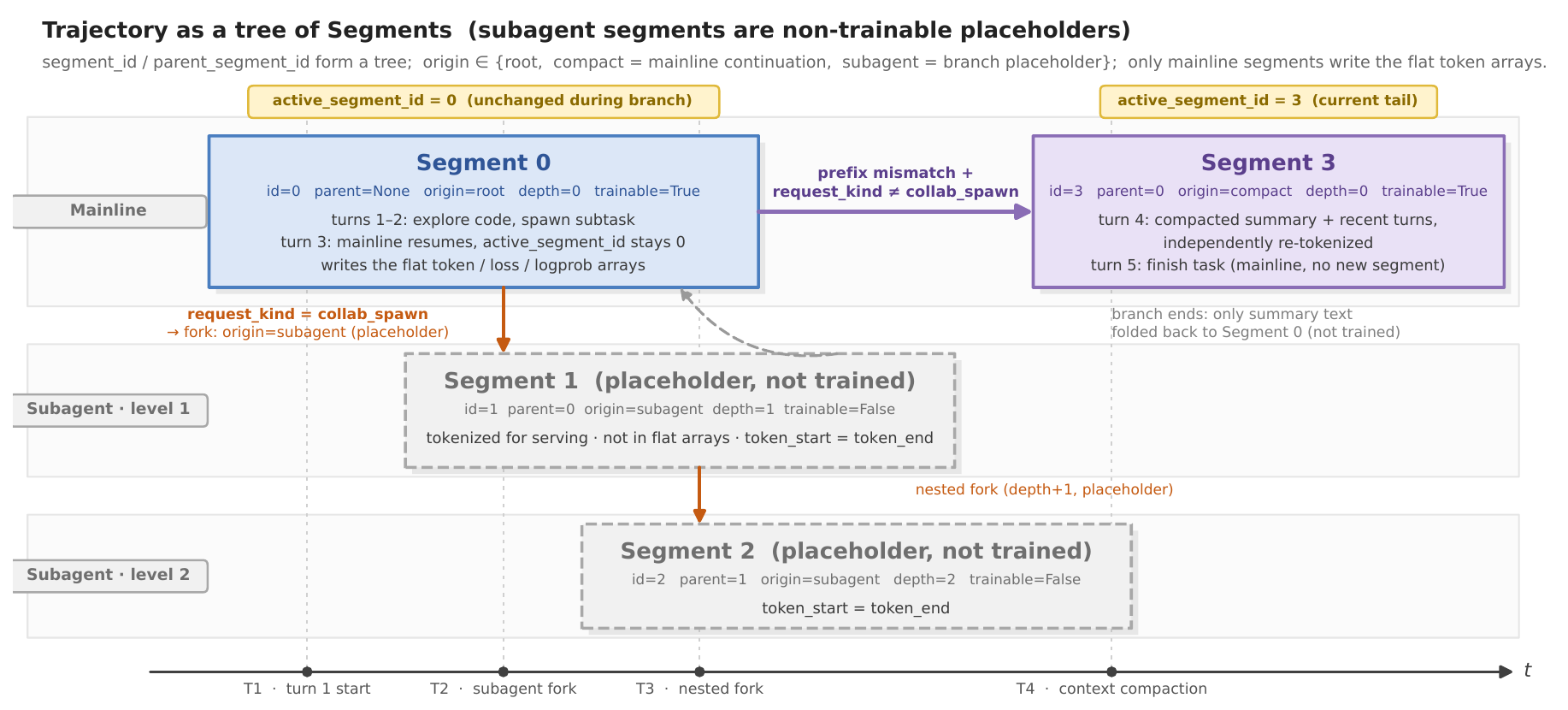}
\caption{AgentFlow stores a black-box trajectory as a tree of segments: trainable mainline segments (\texttt{origin=root} or \texttt{compact}) own the flat token arrays, while a \texttt{collab\_spawn} request forks a non-trainable \texttt{subagent} placeholder (\texttt{origin=subagent}).}
\label{fig:feature-tree}
\end{figure}

\textbf{Compaction and Multi-Agent Control.} Recent black-box agents (e.g., OpenCode, Claude Code, and Codex) natively embed automatic context compaction and multi-agent collaboration mechanisms. These behaviors inherently rewrite or fork the conversation history, violating the standard linear Markov Decision Process (MDP) assumption. Consequently, a linear trajectory store fails prefix matching upon context compaction or subtask delegation. To overcome this, AgentFlow records every trajectory as a directed tree of segments (Figure~\ref{fig:feature-tree}). Tree edges are defined by \texttt{segment\_id} and \texttt{parent\_segment\_id} pointers, while an \texttt{origin} tag explicitly categorizes each node as a mainline start (\texttt{root}), a compaction-induced continuation (\texttt{compact}), or a spawned subtask (\texttt{subagent}).

Crucially, only mainline segments are designated as trainable, earning the right to populate the flat token, loss-mask, and log-probability arrays depicted in Figure~\ref{fig:feature-agentflow}. If a prefix mismatch occurs and the request is not explicitly a \texttt{collab\_spawn}, the trajectory opens a new mainline branch to re-tokenize the compacted context. Conversely, a \texttt{collab\_spawn} request forks a subagent segment one level deeper in the hierarchy. Because multi-agent credit assignment remains an open research problem, LoongSage currently treats subagent segments as non-trainable placeholders that merely track their tree position (\texttt{token\_start = token\_end}). This avoids committing to a reward decomposition prematurely and leaves an extension point for multi-agent RL. In the current parser, every subagent segment is attached to the active mainline segment and therefore sits at depth 1; the nested fork in Figure~\ref{fig:feature-tree} illustrates what the tree format can represent, not a path the current parser produces.

\textbf{Stateful Routing and Protocol Isolation.} This abstraction is enforced via a strict protocol boundary. Upstream parser middleware translates incoming JSON messages into raw token IDs, and symmetrically converts generated token IDs back into the agent's wire format. To maximize inference throughput, the routing layer utilizes session affinity, pinning all requests from a single trajectory to a specific inference worker. This enables KV-cache and prefix reuse across multiple turns, falling back to the least-loaded worker only if the pinned worker has been excluded from service or its in-flight load exceeds a configured threshold.

\begin{figure}[htbp]
\centering
\includegraphics[width=0.7\linewidth]{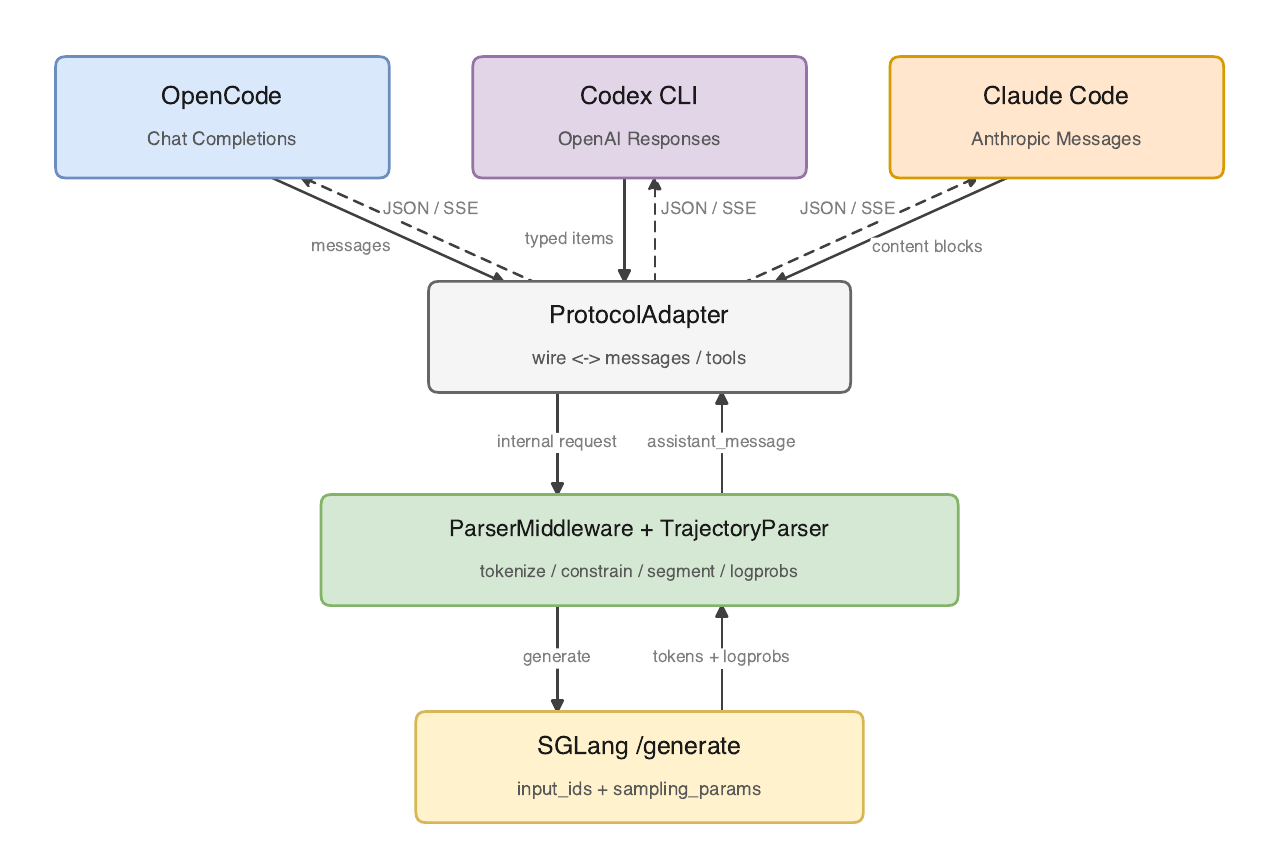}
\caption{The protocol adapter normalizes each agent's wire format into one internal request: OpenCode over Chat Completions, Codex over OpenAI Responses, and Claude Code over Anthropic Messages. A single parser, tokenizer, and SGLang path then serve all of them.}
\label{fig:agent-adapter}
\end{figure}

Black-box agents utilizing proprietary APIs (i.e., those deviating from \texttt{/v1/chat/completions}) are bridged into the unified tree store via protocol adapters (Figure~\ref{fig:agent-adapter}). For instance, Codex (\texttt{/v1/responses}) and Claude Code (\texttt{/v1/messages}) each map to a dedicated adapter that normalizes their wire formats into LoongSage's internal message and tool schemas, keeping the downstream parser, tokenizer, and trajectory storage decoupled from upstream API idiosyncrasies.

\subsection{Staleness-Controlled Fully Asynchronous Training}
\label{sec:async}

In fully asynchronous mode, rollout freshness is no longer implicitly guaranteed by a synchronized schedule. Explicit staleness control thus becomes mandatory when overlapping the training and inference engines to eliminate idle time and repeated initialization overhead.

This producer-consumer architecture aligns with contemporary asynchronous RL systems~\citep{areal,verl,slime,forge}. The rollout engine generates trajectories, a collector aggregates the completed ones into training groups, and the training engine consumes them from a shared buffer, a priority queue keyed by dataset order so that the training sequence stays as close to synchronous mode as completion times permit. Conceptually, this decoupling introduces only two degrees of freedom for data ingestion: determining how many new rollouts to dispatch, and deciding which completed groups are eligible for training. Everything else, including the agents, the trajectory representation, and weight transfer, reuses the synchronous-mode implementation, and the two engines synchronize only at weight-update boundaries.

\textbf{Bounding Staleness.} In a fully asynchronous pipeline, maximizing hardware utilization inherently trades off data freshness. If the inference engine significantly outpaces the trainer, the buffer accumulates trajectories generated by outdated policy weights, increasing off-policy drift. Conversely, if the framework enforces excessively tight synchronization to prevent this drift, the trainer idles while waiting for new rollouts, negating the throughput benefits of asynchrony. To control this trade-off, LoongSage introduces three sliding-window scheduling policies. These policies act as dynamic constraints on the dispatch and collection gates, trading throughput against the version gap between the generation policy and the optimization policy (Table~\ref{tab:window-policies}). Furthermore, the partial-rollout ratio and the version span are computed over the data actually consumed during an optimization step, so each policy's effect on off-policy divergence is directly observable in the training logs.

\begin{table}[ht]
\centering
\footnotesize
\setlength{\tabcolsep}{7pt}
\renewcommand{\arraystretch}{1.2}
\caption{The bounds imposed by each sliding-window strategy at the two control points of the asynchronous runtime, and the guarantees they provide.}
\label{tab:window-policies}
\begin{tabular}{@{}lllll@{}}
\toprule
 & \multicolumn{2}{c}{Bound at the control point} & \multicolumn{2}{c}{Resulting guarantee} \\
\cmidrule(lr){2-3} \cmidrule(lr){4-5}
Strategy & Dispatch & Collection & Staleness & Training order \\
\midrule
\texttt{no-window} & $F + Q \le W$ & unconstrained & unbounded & unconstrained \\
\texttt{window-gated} & $s < s_{\min} + W$, $F + Q \le W$ & unconstrained & $\approx S$ steps & unconstrained \\
\texttt{windowed-fifo} & $F + Q \le W$ & $s - s_{\min} < B$ & indirect & FIFO within $B$ \\
\bottomrule
\end{tabular}
\end{table}

\begin{figure}[htbp]
\centering
\includegraphics[width=\linewidth]{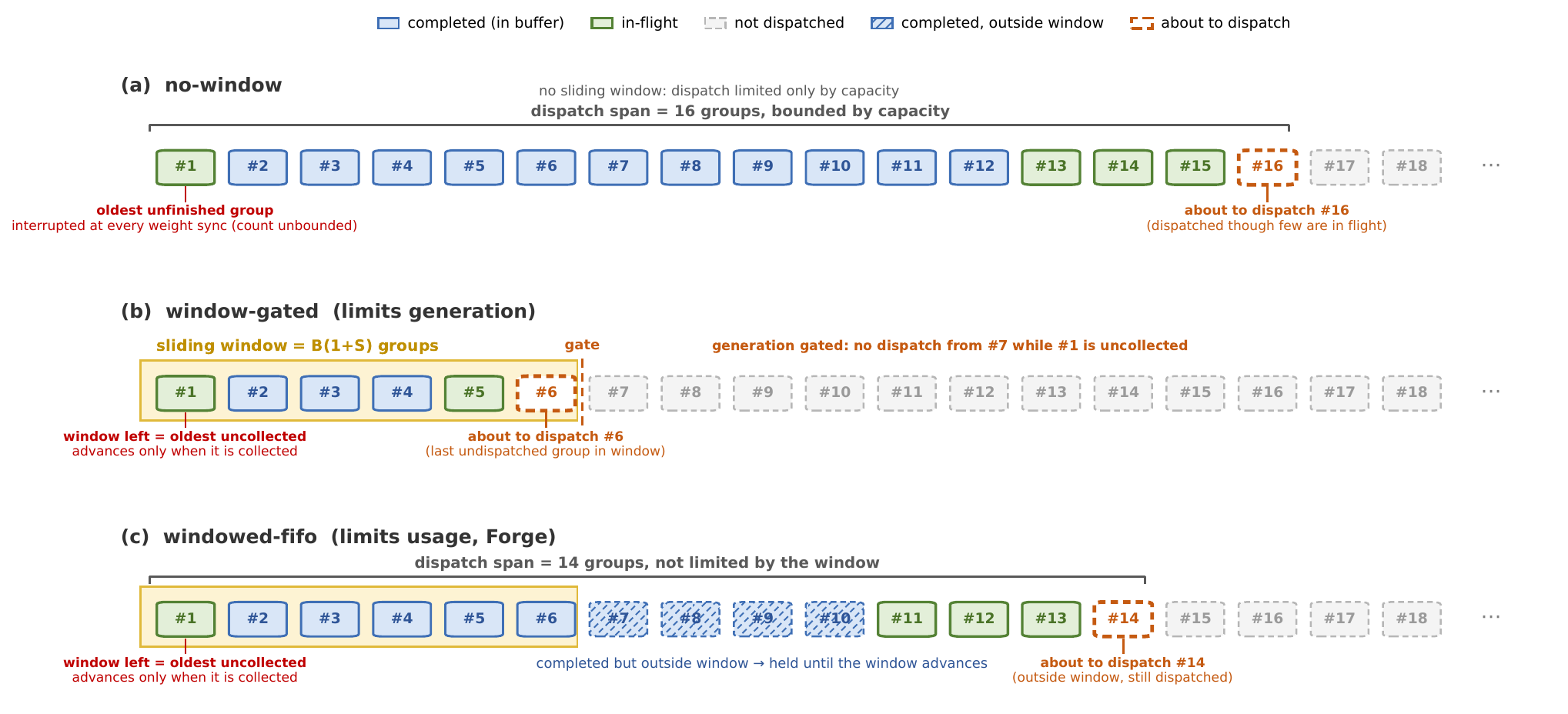}
\caption{The three registered sliding-window policies (a)~\texttt{no-window}, (b)~\texttt{window-gated}, and (c)~\texttt{windowed-fifo} on a shared timeline of dispatched groups (bounds in Table~\ref{tab:window-policies}). The panels are schematic and use different illustrative sizes: the capacity in (a) is $W=16$ groups, the dispatch window in (b) spans $W=6$ groups, the collection window in (c) spans $B=6$ groups, and in (a) and (c) the dispatch span is limited only by the capacity bound $F + Q \le W$, not by any window.}
\label{fig:feature-windows}
\end{figure}

\textbf{Sliding-window policies.} Because a scheduling strategy is entirely defined by the bounds it places on the two control gates, the dispatch gate and the collection gate, the surrounding controller logic remains invariant. Consequently, scheduling is abstracted as a configuration-driven registry seam rather than deeply nested branching logic. Table~\ref{tab:window-policies} formalizes the bounds of the three registered policies, visually mapped onto a shared timeline in Figure~\ref{fig:feature-windows}. 

Throughout this analysis, $B$ denotes \texttt{num\_prompts\_per\_step}, $S$ denotes \texttt{fully\_async.stale\_steps}, and the global window size is $W=B(1+S)$. In the table, $F$ and $Q$ represent the number of groups currently in-flight and buffered, respectively. Variables $s$ and $s_{\min}$ denote the monotonically increasing sequence number of the group in question and that of the oldest dispatched group not yet collected. Notably, in the staleness column, some guarantees hold by strict mathematical construction, whereas others materialize implicitly once uncollected groups accumulate at the collection gate.

\texttt{no-window} (Figure~\ref{fig:feature-windows}a) dispatches a new group as soon as capacity frees up. Its only limit is the capacity bound $F + Q \le W$, which does not bound off-policy drift. Because any completed group is collected into the buffer immediately, short requests finish and enter the buffer in rapid succession. Meanwhile, a computationally heavy, long-horizon request may be interrupted by multiple asynchronous weight updates without bound. This can introduce a length-biased sampling pathology, skewing the training distribution toward shorter trajectories.

\texttt{window-gated} (Figure~\ref{fig:feature-windows}b) imposes a hard gate, refusing to dispatch any task whose sequence number reaches $s_{\min}+W$, i.e., lies $W=B(1+S)$ or more beyond the oldest uncollected one. The window advances only when the oldest group is collected. While this caps the dispatch spread at $W$ groups and thereby bounds staleness to roughly $S$ steps, it risks severe pipeline under-utilization due to \emph{head-of-line blocking} from long-tail requests. In the worst case, the whole pipeline waits for a single straggler to clear the window.

\texttt{windowed-fifo} (Figure~\ref{fig:feature-windows}c) adopts its nomenclature and collection heuristic from Forge. It retains the aggressive capacity-based dispatch rule without generation bounds but restricts collection: a completed group is only ingested if it falls within one batch of the oldest uncollected sequence number. Groups completing too far ahead are held back in the trajectory queue until the window advances. At a weight-update boundary, held complete groups are returned to their data source rather than dropped. This bounds off-policy drift once uncollected groups accumulate and enforces an ordering constraint designed to mitigate the aforementioned short-trajectory bias. A window smaller than one batch would artificially constrain ordering within a single step, while a significantly larger one would simply degenerate into \texttt{no-window}.

The net benefit of asynchronous execution hinges on the relative durations of the rollout and training phases, and the ratio by which the GPU cluster is partitioned between them. Let $x>1$ denote the reciprocal of the cluster fraction dedicated to inference. Here $R_0$ and $T_0$ are the colocated whole-cluster baseline times for rollout and training, $E_0$ is the constant per-step overhead, and $f$ captures the super-linear efficiency of higher per-replica concurrency in a smaller inference pool. Under this parameterization, training time scales as $T_0\,x/(x-1)$ and rollout time as $R_0 f(x)$. Because the two phases overlap, a step's latency is bounded by the slower of the two, which suggests the allocation objective $\min_{x>1}\max\!\left(R_0 f(x),\ T_0\,x/(x-1) + E_0\right)$. The scaling behavior of $f$ is highly workload-dependent, so we use this model as a qualitative guide. Section~\ref{sec:async-experiment} measures the throughput of these policies on one workload.

\subsection{Topology-Aware Weight Synchronization}
\label{sec:transfermesh}

\begin{figure}[ht]
\centering
\includegraphics[width=\linewidth]{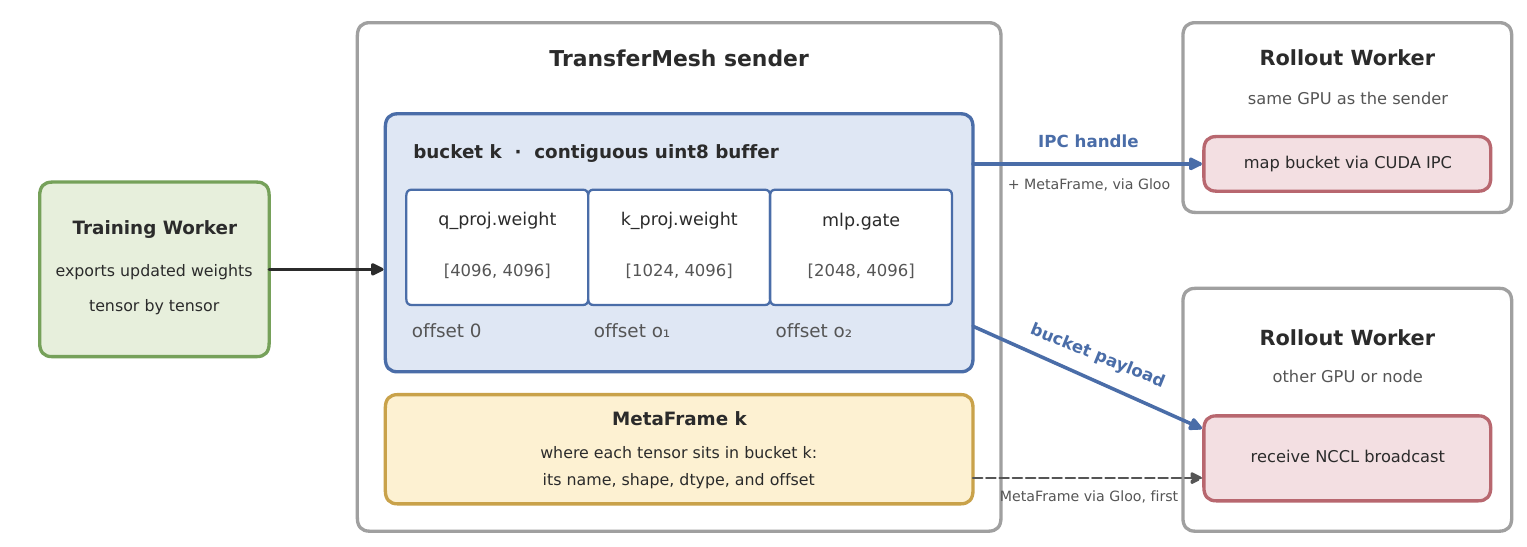}
\caption{TransferMesh packs the named tensor stream into contiguous memory buckets, each described by its own metadata frame, then dynamically selects a transport protocol per receiver based on device topology: utilizing a CUDA IPC handle for colocated engines and an NCCL broadcast for receivers on other GPUs.}
\label{fig:feature-transfermesh}
\end{figure}

TransferMesh provides a high-throughput, engine-agnostic communication fabric for propagating updated parameters from the optimization plane to the inference engine. As training and inference pipelines are increasingly decoupled, weight synchronization can account for a substantial share of per-step overhead for large models. TransferMesh mitigates this by dynamically discovering the physical placement of participating actors during channel initialization and routing traffic across the most efficient hardware-level topology. Figure~\ref{fig:feature-transfermesh} depicts this topology-aware routing architecture.

At its core, TransferMesh abstracts a global weight update as a \emph{named tensor stream}. When the sender yields parameters, the transfer layer coalesces them into contiguous memory buckets and describes each bucket with its own metadata frame recording the tensor name, shape, data type, and byte offset of every tensor it holds. Receivers use these frames to slice each bucket back into named tensors and reconstruct the tensor stream for the inference engine. This decouples the two engines' internal memory layouts and amortizes the high latency and kernel-launch overheads of point-to-point transfers of fragmented, fine-grained tensors.

The underlying transport dynamically adapts to the discovered hardware topology. When training and inference processes are colocated on the same physical device, TransferMesh sends a CUDA Inter-Process Communication (IPC) handle pointing to the sender's memory bucket, together with that bucket's metadata frame. The receiver maps the bucket directly, so the bulk payload is shared without any PCIe/NVLink traversal. For receivers on other GPUs, whether on the same node or a remote one, the bulk payload is sent with an NCCL broadcast; the metadata frame is broadcast first, letting receivers allocate a device buffer of the right size before the payload arrives. All metadata and IPC handles are transmitted out-of-band via lightweight Gloo channels, and receivers restore each parameter with its original shape and precision from the frame.

TransferMesh builds its own process groups and does not share communicators with the backend engines, so weight transfer cannot interleave with their internal data-parallel or tensor-parallel collectives. While the transfer protocol itself is strictly backend-agnostic at the tensor-stream boundary, the currently shipped adapters natively interface with Megatron's parameter export and SGLang's weight-loading routines. Consequently, migrating to an alternative training or inference backend reduces entirely to implementing a localized memory-layout adapter, without modifying the underlying synchronization infrastructure.

\subsection{Full-Vocabulary Multi-Teacher On-Policy Distillation}
\label{sec:opd}

On-policy distillation trains a student model on its self-generated rollouts while leveraging one or more teacher models for dense, token-level distributional supervision. LoongSage implements it as an objective inside the main training loop that can run standalone or combined with algorithms like GRPO or GSPO.

The teacher signal provides two distinct supervision pathways: a policy-gradient regularizer (via a KL penalty on the advantage) and a direct supervised loss interpolated with the RL objective:
\begingroup
\setlength{\abovedisplayskip}{4pt}
\setlength{\abovedisplayshortskip}{4pt}
\setlength{\belowdisplayskip}{4pt}
\setlength{\belowdisplayshortskip}{4pt}
\[
\begin{aligned}
\hat{A}_t
  &= A_t - \lambda_{pg}\,
     D_{\mathrm{KL}}(\pi_{\text{student}} \,\|\, \pi_{\text{teacher}})_t,\\
L &= (1-\lambda_{gkd})L_{RL} + \lambda_{gkd}L_{GKD}.
\end{aligned}
\]
\endgroup
By independently modulating these two pathways---the KL penalty ($\lambda_{pg}$) and the supervised distillation loss ($\lambda_{gkd}$)---the unified loop natively supports PG-style regularization, GKD-style supervised distillation, and hybrid settings. The KL divergence estimator is equally flexible, accommodating sampled-token signals, Top-$K$ truncation, full-vocabulary alignments, and Jensen--Shannon variants based on the available teacher metadata.

\textbf{Teacher Manager and Resource Multiplexing.} Within the multi-teacher paradigm, the Teacher Manager treats teacher models as dynamically scheduled compute resources rather than black-box external API services. Under constrained device memory, multiple teachers are multiplexed onto a single worker group, and inactive teacher weights are offloaded to host CPU memory. Consequently, the permissible number of teachers is bounded by abundant host memory rather than scarce GPU VRAM. The scheduler further guarantees that each teacher is swapped into device memory at most once per forward pass, minimizing PCIe bandwidth thrashing.

\begin{figure}[ht]
\centering
\includegraphics[width=0.8\linewidth]{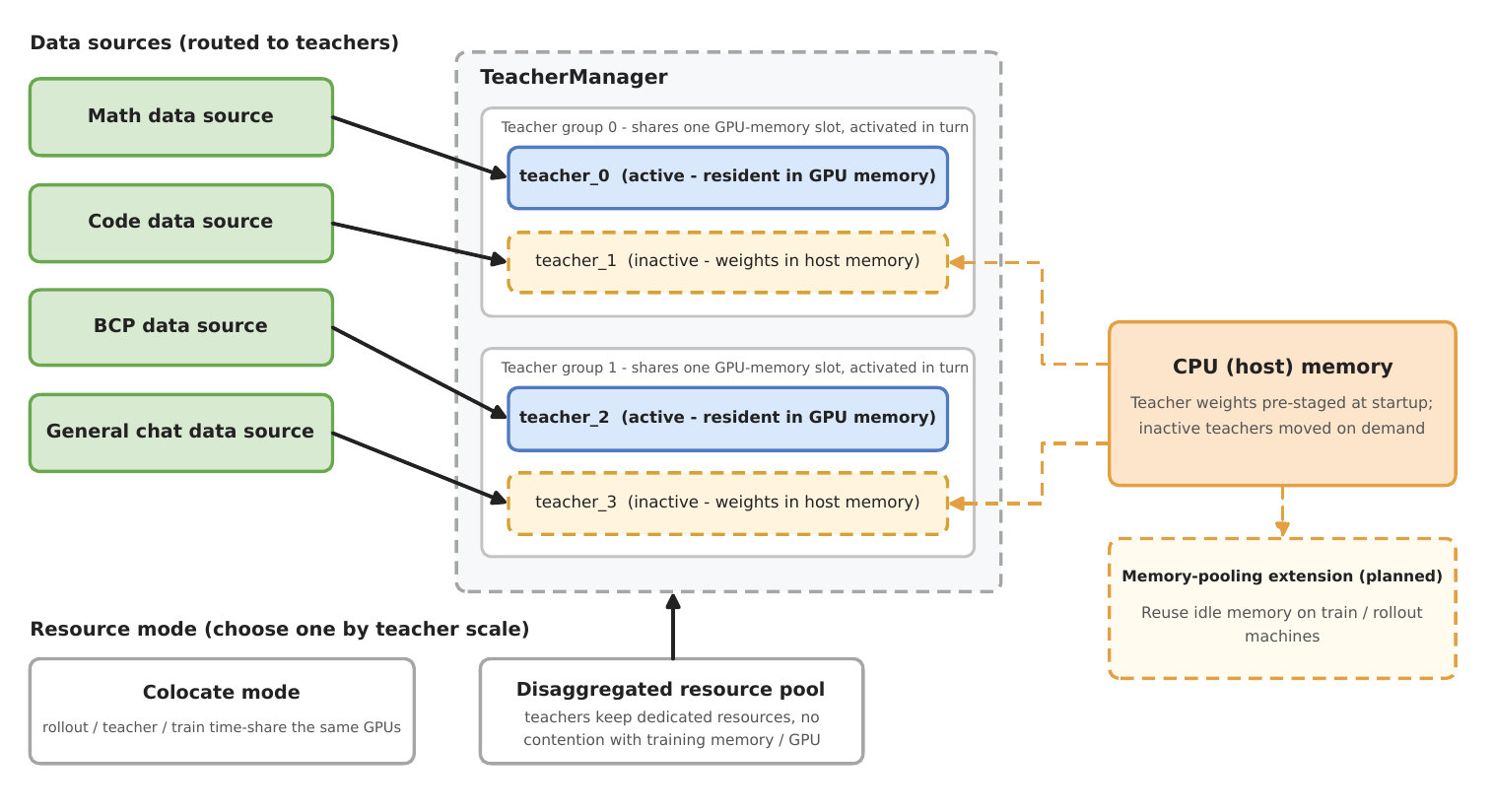}
\caption{The Teacher Manager schedules teacher groups that share a GPU-memory slot, offloading inactive teacher weights to host memory and swapping them back on demand. BCP denotes BrowseComp-Plus.}
\label{fig:mopd-teacher-manager}
\end{figure}

Building on the per-source data plane (Section~\ref{sec:framework}), each data source additionally binds an assigned teacher. Because that plane holds the global data mixture invariant across ranks and mini-batches, the data-mixing ratio maps directly onto each teacher's relative gradient contribution. This architectural choice lets operators steer teacher dominance through data-routing configurations rather than hand-tuned loss-weighting heuristics. Figure~\ref{fig:mopd-teacher-manager} illustrates this resource pooling arrangement. Finally, teacher forward passes execute on the same Megatron backend as the student; this structural symmetry prevents inference-engine numerical discrepancies from corrupting the reverse-KL objective.

\begin{figure}[htbp]
\centering
\includegraphics[width=\linewidth]{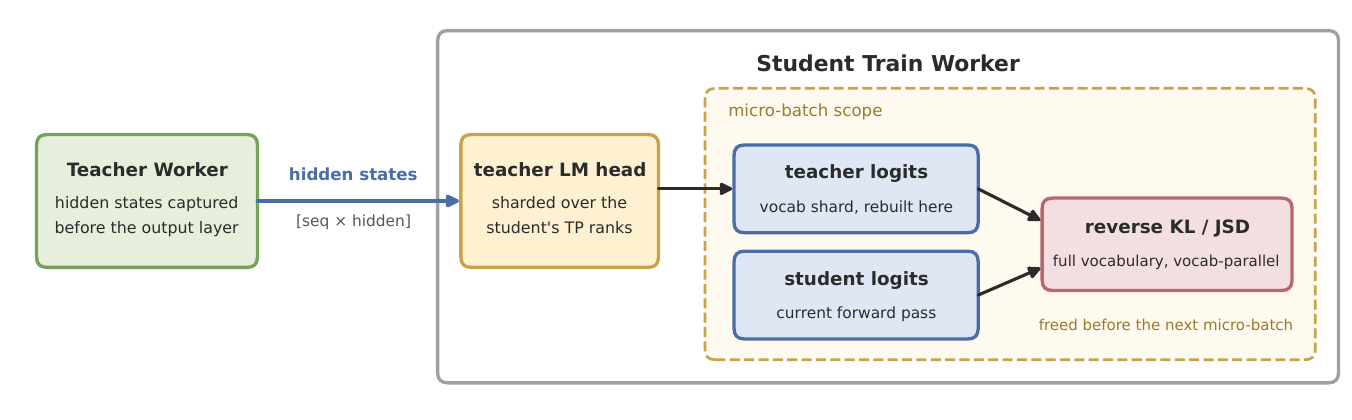}
\caption{Full-vocabulary distillation via micro-batch logit reconstruction: within each micro-batch the teacher logits are rebuilt from hidden states, consumed by the loss, and released before the next micro-batch begins.}
\label{fig:mopd-fullvocab}
\end{figure}

\textbf{Memory-Efficient Full-Vocabulary Distillation.} To avoid the activation-memory bottleneck typically associated with massive vocabularies, LoongSage eschews transferring full probability vectors across nodes. Instead, for the full-vocabulary estimators it transmits compact teacher hidden states---an optimization it shares with DeepSeek-V4~\citep{deepseekv4}---but diverges significantly in its logit materialization strategy. Rather than materializing the full-vocabulary logit tensor for the entire batch, LoongSage shards the teacher's output head using the student's tensor-parallel (TP) layout and reconstructs the required distribution \emph{strictly} at micro-batch granularity. The reconstructed logits are consumed on-the-fly by the loss function and immediately freed before the next micro-batch begins. In the reported single-node, eight-GPU Qwen3-30B-A3B run, full-vocabulary multi-teacher distillation increased peak device memory by only about 17\,GB per GPU relative to DAPO training of the same student on the same node with the same parallel layout (TP${=}4$, PP${=}2$, EP${=}4$). Figure~\ref{fig:mopd-fullvocab} illustrates this reconstruction architecture.

\subsection{Systemic Train--Inference Consistency}
\label{sec:consistency}

\begin{figure}[htbp]
\centering
\includegraphics[width=\linewidth]{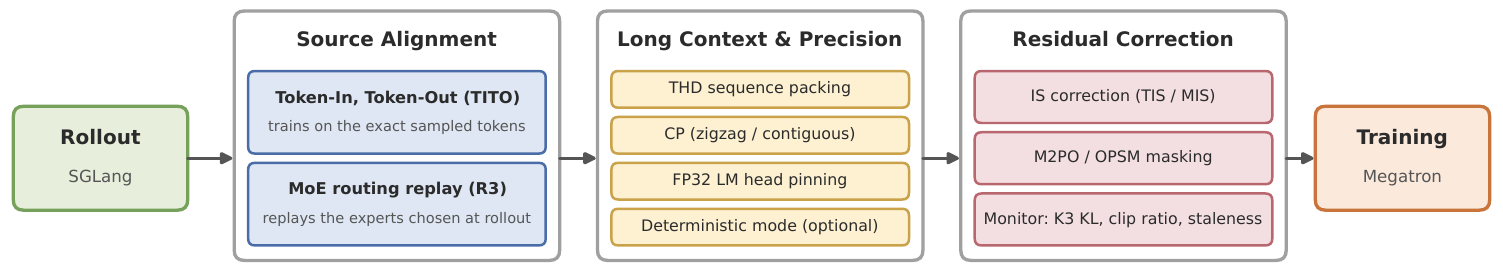}
\caption{Train--inference consistency is handled in three groups: source alignment ensures that training consumes exactly the sampled tokens and expert routes, long-context and precision settings anchor numerical stability, and residual corrections continuously bound the remaining gap.}
\label{fig:feature-consistency}
\end{figure}

On-policy RL fundamentally dictates that the inference and training engines yield nearly identical probability distributions for a given policy state. Any systemic deviation skews the importance-sampling ratio, silently degrading the optimization process into an unstable off-policy regime. Potential sources of divergence span multiple layers: textual re-tokenization mismatches, arithmetic precision truncation, non-equivalent fused kernels~\citep{flashattention}, mismatched quantization schemes between engines, non-deterministic discrete decisions (e.g., Mixture-of-Experts routing), and minor variances in parallel reduction or padding. To counteract this drift, the framework implements a defense-in-depth strategy categorized into source alignment, precision anchoring, and residual correction (summarized in Figure~\ref{fig:feature-consistency}).

\textbf{Deterministic Source Alignment and Routing Replay.} As established in Section~\ref{sec:agentflow}, TITO removes re-tokenization discrepancies at the input boundary. For Mixture-of-Experts (MoE) architectures~\citep{switchtransformer,mixtral,qwen3}, LoongSage integrates Rollout Routing Replay (R3)~\citep{r3}. Because small logit differences can flip expert choices between the two engines, R3 records the exact experts selected by the inference engine for every token across every MoE layer. The framework then deterministically replays this exact routing tensor during the training forward pass. While the router weights continue to receive gradients normally, the computation path is fixed to the sampled trajectory. Replay remains valid under partial rollout, as a trajectory spanning multiple weight versions replays the historical routing context utilized during its original generation.

\textbf{Context Parallelism and Selective Precision Pinning.} To support the long trajectories of agentic tasks without padding artifacts, sequences are tightly packed into THD format and partitioned by Context Parallelism (CP)~\citep{ringattention}, using a zigzag layout by default or contiguous chunks where the model requires it (the DeepSeek-V4 family); MoE routing replay slices the recorded experts with the same layout and therefore remains valid in both cases. While the bulk of the model's compute operates in BF16, LoongSage introduces \emph{Selective Precision Pinning}, allowing developers to anchor highly sensitive parameters to FP32 in both memory storage and layer-wise computation. Because reduced-precision rounding at the output layer distorts softmax log-probabilities, pinning it removes a known source of the gap at a small memory and compute cost. An optional deterministic mode further fixes the training-side reduction order (deterministic Megatron kernels, ring-only NCCL collectives, and deterministic PyTorch/cuDNN algorithms), trading throughput for bitwise-reproducible training passes.

\textbf{Algorithmic Residual Correction and Monitoring.} Because source alignment cannot fully eliminate hardware-level kernel discrepancies or the weight-version staleness inherent to asynchronous execution, the algorithm plane provisions a spectrum of residual defenses spanning both token and sequence granularities. At the fine-grained token level, LoongSage natively supports Truncated Importance Sampling (TIS)~\citep{tis,offpolicyrlframework} and Masked Importance Sampling (MIS). These methods reweight the step-wise loss using the ratio $\pi_{\text{train}}/\pi_{\text{rollout}}$---the discrepancy between the training engine's recomputed probabilities and the inference engine's sampling probabilities---while clipping or bounding out-of-range values.

To mitigate compounded drift over long-horizon tasks, the framework elevates these defenses via explicit sequence-level corrections, which recalibrate the objective based on the cumulative trajectory ratio. Furthermore, it integrates two outlier filters. Second-Moment Trust Policy Optimization (M2PO)~\citep{m2po} masks the tokens with the largest squared log-ratio until the second moment over each data-parallel rank's rollout shard falls below a threshold. Off-Policy Sequence Masking (OPSM)~\citep{deepseekv3.2} drops the gradient of negative-advantage tokens once a sequence's KL exceeds a threshold. Under GRPO, where all tokens share one advantage, the whole sequence then contributes no gradient. Both filters prevent a few highly off-policy tokens or sequences from dominating the gradient. An alternative execution mode treats $\pi_{\text{rollout}}$ as the definitive behavior policy, implicitly absorbing the gap and bypassing old log-probability recomputation---an optimization that is structurally mutually exclusive with TIS, MIS, and M2PO.

The framework continuously monitors the K3 KL divergence estimate, the token-clipping ratio, and weight-version staleness spans. While algorithmic clipping and OPSM act as safeguards against off-policy drift, LoongSage treats aligning critical kernels, replaying discrete MoE decisions, and pinning precision on critical paths as the primary systemic remedies.

\subsection{Lightweight, Usable, and Highly Extensible}
\label{sec:lightweight}

\textbf{Compact Core and Zero-Fork Integration.} After excluding blank lines, comments, and tests, the framework comprises roughly 16,000 lines of Python. The architectural boundaries within LoongSage are designed by applying strict interface segregation. A prime example is TransferMesh (Section~\ref{sec:transfermesh}): the controller builds a single channel descriptor from which both sides derive their endpoints, shielding the training and inference engines from the underlying transport and channel-rebuild complexities. LoongSage enforces a zero-fork dependency model on Megatron-Core, Megatron-Bridge, and SGLang: it pins upstream versions (release tags for SGLang and Megatron-Bridge, a fixed commit for Megatron-LM) instead of maintaining forked repositories. Any transient backend fixes are maintained as isolated patch files applied on top of the pinned versions (plus a small set of runtime monkey-patches for Megatron-Core's distributed optimizer, Transformer Engine's FusedAdam, the CSA CuTe launch cache, and FlashAttention-4's causal-window handling), which can be cleanly deprecated once resolved upstream. By relying on these foundational libraries alongside Ray and Hydra, the framework inherits upstream distributed optimizations and newly supported Megatron-Bridge models with minimal adaptation.

\textbf{Declarative Execution and Validated Recipes.} To maximize usability, LoongSage unifies synchronous, fully asynchronous, multi-teacher distillation, and supervised fine-tuning (SFT) modes behind a single, declarative entry point. Researchers drive complex distributed jobs solely by specifying a configuration recipe. The repository ships with a suite of validated, production-ready recipes---spanning models such as Qwen3-4B, Qwen3-30B-A3B, Qwen3-Coder-30B-A3B, DeepSeek-V4-Flash, and DeepSeek-V4.1-Flash across heterogeneous hardware topologies, including the exact configurations used for the training runs in Section~\ref{sec:experiments}.

\textbf{Configuration-Driven Hot-Plugging.} LoongSage exposes a plugin system for algorithmic iteration. Table~\ref{tab:registry} enumerates the ten explicitly defined core registries. Rather than deeply coupling implementations via rigid subclassing or injecting raw import paths into configuration files, LoongSage utilizes decentralized decorator-based registrations. For instance, applying \texttt{@register\_advantage("grpo")} makes the function self-describing as an extension point, registers its configuration alias, and documents the call signature the implementation is expected to follow. Consequently, searching the codebase for a given decorator surfaces the complete registry of available implementations.

\begin{table}[htbp]
\centering
\scriptsize
\setlength{\tabcolsep}{4pt}
\renewcommand{\arraystretch}{1.25}
\caption{The core registries exposed by LoongSage. Each is invoked via a \texttt{@register\_<name>} decorator at the definition site and bound to the corresponding configuration field. Fields lacking a top-level section (\texttt{agent}, \texttt{reward}, \texttt{sandbox}, \texttt{dataset.*}) are resolved relative to the specific \texttt{data\_source}. The \texttt{kl\_policy} row abbreviates \texttt{opd.pg\_kl\_method} and \texttt{opd.gkd\_kl\_method}.}
\label{tab:registry}
\begin{tabular}{@{}p{.29\linewidth}p{.38\linewidth}p{.27\linewidth}@{}}
\toprule
Registry & Configuration field & Scope \\
\midrule
\texttt{agent} & \texttt{data\_source.agent.name} & Agent loop \\
\texttt{reward} & \texttt{data\_source.reward.name} & Reward computation \\
\texttt{sandbox} & \texttt{data\_source.\allowbreak sandbox.type} & Execution environment \\
\texttt{advantage} & \texttt{algorithm.\allowbreak advantage\_estimator} & Advantage estimation \\
\texttt{policy\_loss} & \texttt{algorithm.\allowbreak policy\_loss} & Policy loss \\
\texttt{kl\_policy} & \texttt{opd.\{pg,gkd\}\_kl\_method} & KL constraint and estimator \\
\texttt{data\_filter} & \texttt{rollout.\allowbreak filter} & Trajectory admission \\
\texttt{buffer\_replay\_strategy} & \texttt{data\_source.dataset.\allowbreak buffer\_replay\_strategy} & Replay of unconsumed prompts \\
\texttt{data\_pre\_processor} & \texttt{data\_source.dataset.\allowbreak data\_pre\_processor} & Dataset preprocessing \\
\texttt{sliding\_window\_strategy} & \texttt{fully\_async.\allowbreak sliding\_window} & Asynchronous scheduling \\
\bottomrule
\end{tabular}
\end{table}
\section{Experiments}
\label{sec:experiments}

\subsection{Experimental Setup}

To comprehensively evaluate LoongSage across diverse agentic RL tasks and training paradigms, we analyze metrics from four complete training runs, summarized in Table~\ref{tab:experiment-config}. The runs span mathematical reasoning (DAPO), long-horizon software engineering (SWE), black-box coding agents (OpenCode), and multi-teacher on-policy distillation (MOPD). The SWE and OpenCode configurations derive tasks from R2E-Gym~\citep{r2egym}, the DAPO run utilizes DAPO-Math-17k~\citep{dapo}, and the MOPD setup mixes BrowseComp-Plus with DAPO-Math-17k. All runs use the Megatron and SGLang backends described in Section~\ref{sec:framework}. Unless explicitly stated otherwise, each reported metric traces a single, continuous full-lifecycle run rather than an average across multiple short-lived trials.

Our evaluation tracks three primary classes of metrics: training dynamics, train--inference consistency, and end-to-end efficiency. Training entropy reflects the concentration of the policy distribution. \texttt{rollout/reward\_mean} quantifies task performance on sampled trajectories, and \texttt{train/is\_approx\_k3\_kl} measures the approximate probability divergence between the sampling and training phases using the K3 estimator~\citep{klapprox}. \texttt{timing/step} captures the end-to-end latency of a training step. Because agentic trajectories exhibit extreme variance in horizon length and computational complexity, our analysis focuses on macro-level convergence trends and the absence of sustained systemic divergence, rather than localized micro-fluctuations.

\begin{table}[htbp]
\centering
\footnotesize
\setlength{\tabcolsep}{5pt}
\renewcommand{\arraystretch}{1.25}
\caption{The runs reported in this paper.}
\label{tab:experiment-config}
\begin{tabular}{@{}p{.12\linewidth}p{.30\linewidth}l p{.36\linewidth}@{}}
\toprule
Setting & Model & Hardware & Focus \\
\midrule
\multicolumn{4}{c}{\emph{Training runs, one complete run each}} \\
\cmidrule{1-4}
SWE & DeepSeek-V4-Flash & 8 H20 nodes & Long-horizon software tasks \\
DAPO & DeepSeek-V4-Flash & 6 H20 nodes & Verifiable mathematical reasoning \\
OpenCode & Qwen3-Coder-30B-A3B & 4 H20 nodes & Black-box coding agent \\
MOPD & Qwen3-30B-A3B & 1 H20 node & Multi-teacher distillation \\
\addlinespace[2pt]
\cmidrule{1-4}
\addlinespace[2pt]
\multicolumn{4}{c}{\emph{Controlled comparisons between configurations}} \\
\cmidrule{1-4}
Async & Qwen3-30B-A3B & 2 H20 nodes & Asynchronous throughput (Sec.~\ref{sec:async-experiment}) \\
\bottomrule
\end{tabular}
\end{table}

\subsection{Training Results}

\begin{figure}[htbp]
\centering
\includegraphics[width=.78\linewidth]{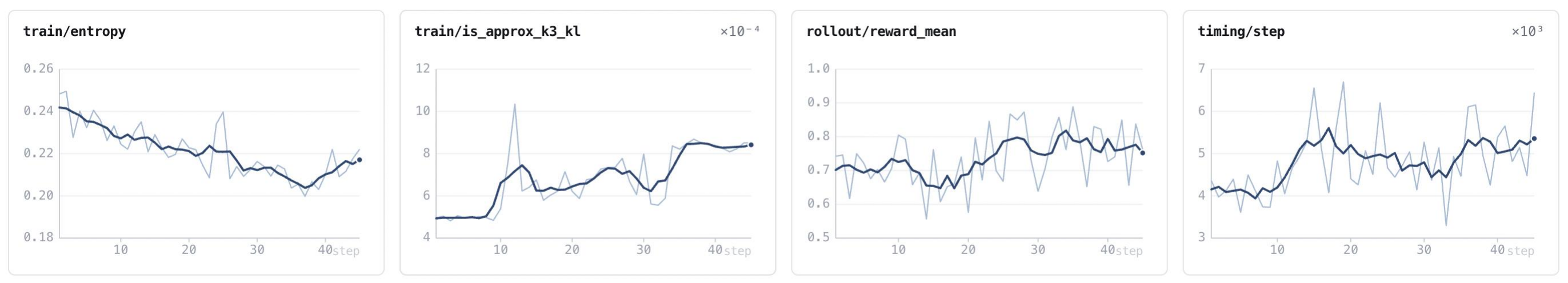}
\caption{Training curves for DeepSeek-V4-Flash on SWE tasks.}
\label{fig:exp-dsv4-swe}
\end{figure}

\paragraph{SWE Tasks.}
In the SWE run (Figure~\ref{fig:exp-dsv4-swe}), the metric that moves most is the train--inference gap. The K3 estimate stays at the $10^{-4}$ scale but rises in two phases, from about $5\times10^{-4}$ to about $8.4\times10^{-4}$ (with a transient spike near $10^{-3}$ early in the run), and holds near that level over the final steps. Over the same period, mean rollout reward climbs and stabilizes in the $0.75$--$0.80$ range, and entropy decreases overall from about $0.24$ to $0.21$, with a slight rebound near the end. Step time grows from about $4100$ to about $5300$ seconds.

\begin{figure}[htbp]
\centering
\includegraphics[width=.78\linewidth]{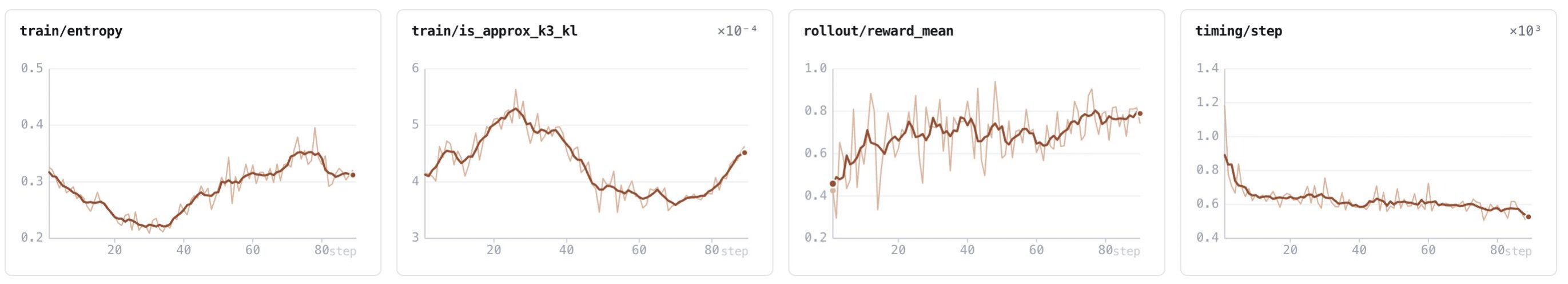}
\caption{Training curves for DeepSeek-V4-Flash on DAPO mathematical reasoning.}
\label{fig:exp-dsv4-dapo}
\end{figure}

\paragraph{DAPO Mathematical Reasoning.}
Entropy in the DAPO run (Figure~\ref{fig:exp-dsv4-dapo}) is not monotonic: it first dips to about $0.22$ and then recovers to about $0.30$, while reward rises from roughly $0.45$ to nearly $0.80$. The K3 gap follows its own arc, peaking at $5.3\times10^{-4}$, declining, and rising again to about $4.5\times10^{-4}$ at the end of the run. Step time falls from $900$ to $550$ seconds, with most of the drop within the first ten steps.

\begin{figure}[htbp]
\centering
\includegraphics[width=.78\linewidth]{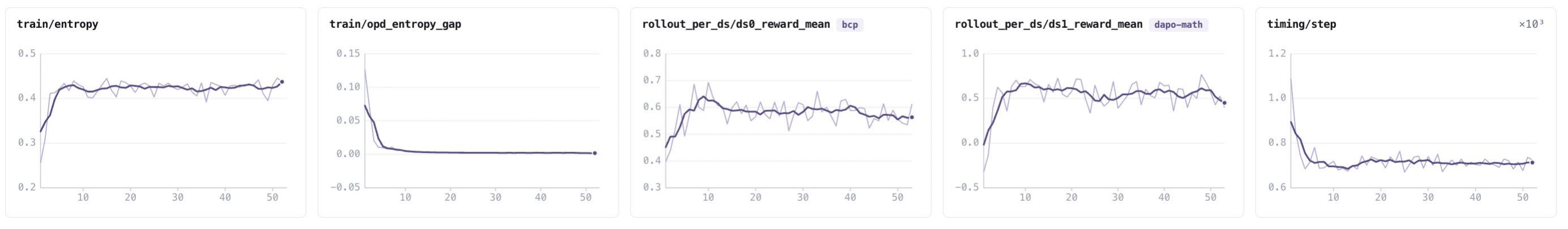}
\caption{Training curves for full-vocabulary multi-teacher MOPD on Qwen3-30B-A3B.}
\label{fig:exp-qwen-mopd}
\end{figure}

\paragraph{Full-Vocabulary Multi-Teacher Distillation.}
The MOPD run uses two teachers trained from Qwen3-30B-A3B with the repository's RL recipes (a BrowseComp-Plus (BCP) expert and a math expert), one per data source at a 1:1 prompt ratio, with pure GKD ($\lambda_{pg}{=}0$, $\lambda_{gkd}{=}1$) and the full-vocabulary KL estimator. Task reward is thus a diagnostic signal rather than the optimization objective, and the run validates the system path rather than a capability gain: teacher multiplexing, micro-batch full-vocabulary logit reconstruction, a bounded memory overhead (Section~\ref{sec:opd}), and stable step time. Unlike the RL runs, whose entropy ends at or below its starting level, distillation raises the student's entropy. In this run (Figure~\ref{fig:exp-qwen-mopd}), the student--teacher entropy gap closes almost immediately, and over the same steps the student's entropy rises from $0.31$ to about $0.42$, where it stays. Rewards on both data sources rise quickly in the early steps and then decline slightly over the rest of the run; this mild drift coincides with the student matching the teachers' entropy. Step time settles from $1100$ to $700$ seconds.

\begin{figure}[htbp]
\centering
\includegraphics[width=.78\linewidth]{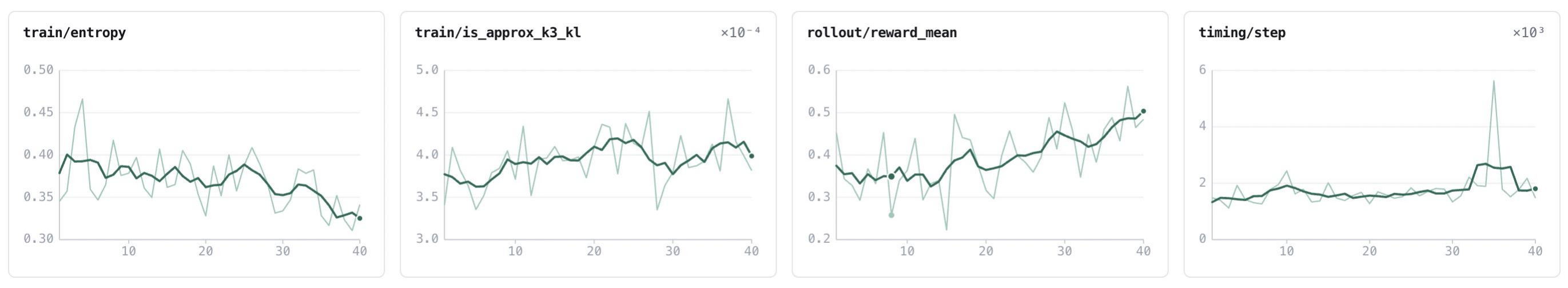}
\caption{Training curves for Qwen3-Coder-30B-A3B with the OpenCode black-box agent.}
\label{fig:exp-qwen-opencode}
\end{figure}

\paragraph{Black-Box OpenCode Agent.}
The OpenCode run (Figure~\ref{fig:exp-qwen-opencode}) trains through an agent whose internals LoongSage does not control. Reward rises from $0.35$ to $0.50$, and the K3 gap stays near $4\times10^{-4}$. Step time mostly stays between about $1300$ and $2000$ seconds, apart from a single step near step $35$ that takes about $5600$ seconds, a long-tail episode typical of black-box coding agents.

\subsection{Asynchronous Throughput}
\label{sec:async-experiment}

While the preceding training runs validate end-to-end algorithmic convergence, they do not isolate the pure system-level throughput gains yielded by asynchronous execution. To quantify this, we execute a controlled ablation on a workload where the long-tail latency is severe enough for pipeline overlap to yield material benefits. 

We benchmark Qwen3-30B-A3B in BF16 with TP${=}4$, PP${=}2$, and EP${=}4$ across two H20 nodes (16 GPUs). The algorithm configuration enforces GRPO with importance-sampling correction, M2PO~\citep{m2po}, and OPSM~\citep{deepseekv3.2}. Prompts are drawn from DAPO-Math-17k, with responses hard-capped at $20$K tokens. Each step samples $B{=}64$ prompts generating $8$ trajectories apiece. The benchmark executes for exactly $100$ steps. The \emph{colocated baseline} forces the trainer and rollout engines to time-multiplex all $16$ GPUs. Conversely, the \emph{fully asynchronous} configurations enforce an $8{:}8$ spatial split, dedicating two 4-GPU SGLang replicas exclusively to inference. All asynchronous runs enable partial rollout and mask the resulting off-policy spans. Because the initial step (Step $0$) inherently incurs the cold-start overheads of prefix-cache warm-up, KV-cache allocation, and initial buffer population, we report both the aggregate $100$-step average and the steady-state average over the subsequent $99$ steps.

Table~\ref{tab:async-throughput} summarizes the throughput gains. The best tested asynchronous configuration reduces the average step time from $580.79$s to $428.76$s over $100$ steps (a $1.35\times$ speedup); excluding the cold-start step, the steady-state average falls to $414.58$s, a $1.40\times$ speedup. Globally, the end-to-end latency reduction across asynchronous configurations ranges from $22\%$ to $26\%$. While partial rollout alone provides a $1.19\times$ boost in colocated mode, the asynchronous \texttt{windowed-fifo} strategy compounds this advantage, accelerating the steady-state step by an additional $13\%$--$15\%$ over the optimized colocated baseline (whereas \texttt{no-window} yields only a $7\%$ marginal gain).

\begin{table}[htbp]
\centering
\footnotesize
\setlength{\tabcolsep}{5pt}
\renewcommand{\arraystretch}{1.25}
\caption{Average step time under colocated and fully asynchronous execution. Speedups are relative to the colocated baseline in the same column.}
\label{tab:async-throughput}
\begin{tabular}{@{}p{.34\linewidth}c rr rr@{}}
\toprule
& & \multicolumn{2}{c}{all $100$ steps} & \multicolumn{2}{c}{excluding step $0$} \\
\cmidrule(lr){3-4}\cmidrule(lr){5-6}
Configuration & Split & avg (s) & speedup & avg (s) & speedup \\
\midrule
Colocated & 16 & 580.79 & 1.00$\times$ & 578.63 & 1.00$\times$ \\
Colocated $+$ partial rollout & 16 & 489.63 & 1.19$\times$ & 486.34 & 1.19$\times$ \\
\addlinespace[2pt]
Async, \texttt{no-window}, $S{=}1.0$ & 8:8 & 454.50 & 1.28$\times$ & 451.74 & 1.28$\times$ \\
Async, \texttt{windowed-fifo}, $S{=}1.0$ & 8:8 & 434.76 & 1.34$\times$ & 424.30 & 1.36$\times$ \\
Async, \texttt{windowed-fifo}, $S{=}2.0$ & 8:8 & 428.76 & 1.35$\times$ & 414.58 & 1.40$\times$ \\
\bottomrule
\end{tabular}
\end{table}

Beyond aggregate latency reductions, the ablation shows two effects. First, at $S{=}1.0$ the \texttt{windowed-fifo} strategy achieves a $4.5\%$ higher throughput than \texttt{no-window} over all $100$ steps ($6.5\%$ excluding step $0$), despite a buffer of $13$ groups versus $27$ at step boundaries. The shallower buffer is consistent with its collection window pacing the consumer against the producer, so that early-finishing groups cannot fill the buffer ahead of older ones. For this workload, the ordering constraint therefore costs no throughput. Second, relaxing the staleness bound $S$ from $1.0$ to $2.0$ elevates the steady-state speedup from $1.36\times$ to $1.40\times$. However, the deeper buffer requires a longer initialization phase. Thus, the $100$-step aggregate speedup barely shifts, and the true end-to-end gain only converges to the steady-state asymptote as the total step count scales.

\section{Conclusion}
\label{sec:conclusion}

In this paper, we introduced LoongSage, a lightweight, agent-native Reinforcement Learning framework engineered specifically for the systemic complexities of LLM post-training. Under its single-controller orchestration model, LoongSage unifies arbitrary agent interactions, multi-turn trajectory materialization, multi-teacher distillation, and distributed policy optimization into a cohesive, modular architecture. 

Built upon this foundation, the framework delivers six architectural contributions: (1) zero-intrusion TITO integration for diverse white- and black-box agent topologies; (2) staleness-controlled, fully asynchronous pipeline execution; (3) topology-aware TransferMesh weight synchronization; (4) memory-multiplexed, full-vocabulary multi-teacher distillation; (5) systemic train--inference consistency maintained by design; and (6) a lightweight, zero-fork core exposing a configuration-driven plugin system. Collectively, these mechanisms address the engineering bottlenecks imposed by long-horizon multi-turn rollouts, severe heavy-tail latency, and large-scale distributed resource coordination. Training runs spanning mathematical reasoning, complex software engineering, black-box coding agents, and multi-teacher distillation show that LoongSage drives agentic training while keeping the train--inference gap of the RL runs at the $10^{-4}$ scale. 

Looking ahead, our roadmap focuses on evolving LoongSage into a comprehensive post-training ecosystem. To further optimize memory utilization at scale, we will introduce cross-node distributed storage for multi-teacher weights and native low-precision training primitives. Algorithmically, we intend to broaden the framework's unified pipeline to encompass Low-Rank Adaptation (LoRA), alongside native support for omni-modal architectures. On the inference side, we plan to complete prefill--decode disaggregation. We hope LoongSage provides the open-source and academic communities with a highly usable, extensible foundation, accelerating future breakthroughs in large-scale agentic AI systems.

\section*{Acknowledgments}
The birth of LoongSage is inseparable from a thriving open-source community, and we extend our special thanks to the excellent projects that made this possible: Megatron-LM and Megatron-Bridge for large-scale distributed training and weight bridging, SGLang for high-performance inference, and Ray for flexible distributed scheduling.

In terms of architecture design and code implementation, LoongSage draws deep inspiration from and heavily references several outstanding works: slime, whose architecture and logic provided significant inspiration for our minimalist design; verl, which offered an excellent design paradigm for an extensible RL training and inference framework; and Agent-Lightning, which inspired our approach to decoupling Agent frameworks from RL post-training platforms.


\bibliography{iclr2025_conference}
\bibliographystyle{iclr2025_conference}

\clearpage
\appendix
\section{Authors}

Authors are listed alphabetically by their first name.

\textbf{Research \& Engineering:} Bowei Song, Caiwei Zhao, Chuwen Liu, Dechuang Gu, Dong Fang, Feiyu Xie, Huaiyu Li, Hui Zhang, Jian Tian, Jian Wang, Jie Zhao, Liguang Li, Maoyun Wang, Meiyan Yang, Minghan Zhu, Pai Liu, Pandeng Yao, Renbao Tian, Ru Yang, Shuai Jia, Shuai Tan, Shuailong Li, Shuliang Li, Teng Jin, Tianze Yang, Wenbo Cui, Wenjing Bai, Wenkang Wu, Xiaowei Jiang, Xiaoyi Dong, Yanpeng Wang, Yanwen Wang, Yaohui Ma, Yawen Wang, Yihang Zhang, Yingjie Yan, Zhanbiao Lian, Zhenyu Lu, Zhida Hu, Zhihui Hao, Zhihui Hu, Zhiwen Xiao, Zhiyi Gao, Zhongwei Qi, Zhuangzhuang Wu, Ziqian Liu, Ziyue Zheng.

\section{Operational Notes and Limitations}
\label{sec:appendix-ops}

Stabilizing the DeepSeek-V4-Flash SWE training recipe surfaced a range of systemic failures that are invisible at the algorithmic level, yet determine whether a large-scale distributed job can reach completion. We categorize these observations to aid future infrastructure developers:

\textbf{Hard Systemic Failures.} These faults abruptly halt execution, manifesting through explicit crash signals or out-of-memory (OOM) triggers. Examples encountered during scale-up include: numerical instability (NaNs) in early Megatron-Core versions when shared-expert overlap was enabled; SGLang exhausting device VRAM specifically during topology weight updates; a cuDNN DSA kernel triggering illegal memory corruption; and a CuTe launch cache silently leaking after several hundred steps. Furthermore, an FP32 output-layer pinning unexpectedly broke Megatron's checkpoint saver, while the CPU optimizer exhausted host memory and crashed. At the framework orchestration level, repeated JIT operator recompilation under dynamic batching induced steady host RAM bloating, and a memory-saver hook conflict between SGLang and Megatron-Core deterministically crashed the inference engine upon CUDA-graph capture. 

\textbf{Silent Structural and Arithmetic Corruptions.} More subtly, these anomalies yield no explicit crash signals, allowing training to proceed while corrupting the RL objective. For instance, a PDL dependency anomaly within a FlashInfer CuTe-DSL \texttt{rmsnorm} kernel introduced an asynchronous timing hazard, silently perturbing normalization statistics and inflating the measured train--inference gap. In another instance, a parameter had identical tensor shapes across both engines but used different memory layouts, causing weight updates to complete without error yet populate the tensor with corrupted data. Similarly, an incorrectly implemented clamp in the shared-expert activation quietly widened the distribution gap. Because such defects fail silently, the policy can degrade for extended periods while conventional metrics appear normal---underscoring the necessity of continuously monitoring the sampling--training gap rather than verifying it only at bring-up.

These real-world challenges highlight a key realization for the agentic RL era: \emph{Reinforcement Learning acts as a strong amplifier for software and hardware coupling}. Minute variances in operator implementations, non-deterministic CUDA rounding, reduction-order variance in collectives, and dynamic batch-size adjustments are fed back into the policy via the reward signal. Consequently, \emph{hyperparameters alone are insufficient to fully describe a reproducible RL recipe}. The configurations shipped with LoongSage encode the resolutions to these systemic hazards, making them production-validated deployment blueprints rather than mere algorithmic settings.

\textbf{Current Architectural Boundaries and Roadmap.} First, for certain black-box agents, the framework's parser cannot yet deterministically classify whether a new branch stems from context compaction or a spawned subtask, necessitating that subtask segments temporarily remain non-trainable placeholders. Second, cross-node memory pooling for teacher weights---designed to utilize idle host RAM on training and rollout nodes as a global distributed weight cache---currently exists as a defined interface awaiting backend implementation. Third, the fully asynchronous mode does not yet match the synchronous path's generality regarding heterogeneous multi-data-source routing and multi-agent topologies, and its asynchronous metric aggregation requires finer precision tuning. 

Taken together, these boundaries feed directly into the broader roadmap outlined in Section~\ref{sec:conclusion}, which additionally targets prefill--decode disaggregation for inference to relieve the operational bottlenecks discussed above. Each limitation is scoped against a stable interface, so closing it does not require changes to the orchestration core.

\end{document}